\PassOptionsToPackage{unicode}{hyperref}
\PassOptionsToPackage{hyphens}{url}
\PassOptionsToPackage{dvipsnames,svgnames,x11names}{xcolor}
\documentclass[
  letterpaper,
]{article}
\usepackage{xcolor}
\usepackage[margin=1in]{geometry}
\usepackage{amsmath,amssymb}
\usepackage{iftex}
\ifPDFTeX
  \usepackage[T1]{fontenc}
  \usepackage{textcomp} % provide euro and other symbols
\else % if luatex or xetex
  \usepackage{unicode-math} % this also loads fontspec
  \defaultfontfeatures{Scale=MatchLowercase}
  \defaultfontfeatures[\rmfamily]{Ligatures=TeX,Scale=1}
\fi
\usepackage{lmodern}
\ifPDFTeX\else
\fi
\IfFileExists{upquote.sty}{\usepackage{upquote}}{}
\IfFileExists{microtype.sty}{% use microtype if available
  \usepackage[]{microtype}
  \UseMicrotypeSet[protrusion]{basicmath} % disable protrusion for tt fonts
}{}
\makeatletter
\@ifundefined{KOMAClassName}{% if non-KOMA class
  \IfFileExists{parskip.sty}{%
    \usepackage{parskip}
  }{% else
    \setlength{\parindent}{0pt}
    \setlength{\parskip}{6pt plus 2pt minus 1pt}}
}{% if KOMA class
  \KOMAoptions{parskip=half}}
\makeatother
\makeatletter
\ifx\paragraph\undefined\else
  \let\oldparagraph\paragraph
  \renewcommand{\paragraph}{
    \@ifstar
      \xxxParagraphStar
      \xxxParagraphNoStar
  }
  \newcommand{\xxxParagraphStar}[1]{\oldparagraph*{#1}\mbox{}}
  \newcommand{\xxxParagraphNoStar}[1]{\oldparagraph{#1}\mbox{}}
\fi
\ifx\subparagraph\undefined\else
  \let\oldsubparagraph\subparagraph
  \renewcommand{\subparagraph}{
    \@ifstar
      \xxxSubParagraphStar
      \xxxSubParagraphNoStar
  }
  \newcommand{\xxxSubParagraphStar}[1]{\oldsubparagraph*{#1}\mbox{}}
  \newcommand{\xxxSubParagraphNoStar}[1]{\oldsubparagraph{#1}\mbox{}}
\fi
\makeatother

\usepackage{longtable,booktabs,array}
\usepackage{calc} % for calculating minipage widths
\usepackage{etoolbox}
\makeatletter
\patchcmd\longtable{\par}{\if@noskipsec\mbox{}\fi\par}{}{}
\makeatother
\IfFileExists{footnotehyper.sty}{\usepackage{footnotehyper}}{\usepackage{footnote}}
\makesavenoteenv{longtable}
\usepackage{graphicx}
\makeatletter
\newsavebox\pandoc@box
\newcommand*\pandocbounded[1]{% scales image to fit in text height/width
  \sbox\pandoc@box{#1}%
  \Gscale@div\@tempa{\textheight}{\dimexpr\ht\pandoc@box+\dp\pandoc@box\relax}%
  \Gscale@div\@tempb{\linewidth}{\wd\pandoc@box}%
  \ifdim\@tempb\p@<\@tempa\p@\let\@tempa\@tempb\fi% select the smaller of both
  \ifdim\@tempa\p@<\p@\scalebox{\@tempa}{\usebox\pandoc@box}%
  \else\usebox{\pandoc@box}%
  \fi%
}
\def\fps@figure{htbp}
\makeatother

\providecommand{\tightlist}{%
  \setlength{\itemsep}{0pt}\setlength{\parskip}{0pt}}

\usepackage[]{natbib}
\DeclareUnicodeCharacter{2212}{\ensuremath{-}}
\DeclareUnicodeCharacter{2013}{--}
\DeclareUnicodeCharacter{2014}{---}
\DeclareUnicodeCharacter{2009}{\,}
\DeclareUnicodeCharacter{00A0}{~}
\DeclareUnicodeCharacter{00D7}{\ensuremath{\times}}
\DeclareUnicodeCharacter{2264}{\ensuremath{\leq}}
\DeclareUnicodeCharacter{2265}{\ensuremath{\geq}}
\DeclareUnicodeCharacter{2260}{\ensuremath{\neq}}
\DeclareUnicodeCharacter{2248}{\ensuremath{\approx}}
\DeclareUnicodeCharacter{03B1}{\ensuremath{\alpha}}
\DeclareUnicodeCharacter{03B2}{\ensuremath{\beta}}
\DeclareUnicodeCharacter{03C3}{\ensuremath{\sigma}}
\DeclareUnicodeCharacter{03C7}{\ensuremath{\chi}}
\DeclareUnicodeCharacter{2032}{\ensuremath{'}}

\usepackage{caption}
\usepackage{multirow}

\setcitestyle{round}

\AtBeginEnvironment{apptbl}{\footnotesize}
\makeatletter
\newcommand\table@defaultsize{\footnotesize}
\let\table@size\table@defaultsize
\newcommand\nexttablesize[1]{\gdef\table@size{#1}}
\AtBeginEnvironment{longtable}{\table@size\global\let\table@size\table@defaultsize}
\makeatother

\makeatletter
\let\floatnote@text\@empty
\newcommand\floatnote[1]{\gdef\floatnote@text{#1}}
\newcommand\fs@plainnote{\fs@plain
  \def\@fs@post{\ifx\floatnote@text\@empty\else
    \vspace\abovecaptionskip\noindent\floatnote@text\par
    \global\let\floatnote@text\@empty\fi}}
\makeatother

\AtBeginDocument{%
  \floatstyle{plainnote}\restylefloat{appfig}%
  \floatplacement{appfig}{H}\floatplacement{apptbl}{H}}

\AtBeginDocument{}
\makeatletter
\renewcommand\@maketitle{%
  \newpage\null\vskip 0.5em
  \begin{center}
    {\Large\bfseries \@title\par}
    \vskip 1.5em
    {\normalsize \@author\par}
  \end{center}
  \par\vskip 1em}
\newcommand\aclauthor[3]{% name, affiliation lines, email
  \begin{tabular}[t]{@{}c@{}}
    \textbf{#1}\\[1pt]
    #2\\[1pt]
    {\small\texttt{#3}}
  \end{tabular}}
\makeatother

\AtBeginDocument{%
  \author{%
    \aclauthor{Davood Wadi}
      {Desautels Faculty of Management\\ McGill University\\ Montreal, Quebec, Canada}
      {davood.wadi@hec.ca}%
    \hspace{3em}%
    \aclauthor{Yu Ma}
      {Desautels Faculty of Management\\ McGill University\\ Montreal, Quebec, Canada}
      {yu.ma@mcgill.ca}%
  }%
}
\makeatletter
\@ifpackageloaded{float}{}{\usepackage{float}}
\floatstyle{plain}
\@ifundefined{c@chapter}{\newfloat{apptbl}{h}{loapptbl}}{\newfloat{apptbl}{h}{loapptbl}[chapter]}
\floatname{apptbl}{Table W}
\floatstyle{plaintop}
\restylefloat{apptbl}
\newcommand*\quartoapptblref[1]{Table \hyperref[#1]{W\ref{#1}}}
\@ifpackageloaded{caption}{}{\usepackage{caption}}
\DeclareCaptionLabelFormat{quartoapptblreflabelformat}{#1#2}
\makeatother
\makeatletter
\@ifpackageloaded{float}{}{\usepackage{float}}
\floatstyle{plain}
\@ifundefined{c@chapter}{\newfloat{appfig}{h}{loappfig}}{\newfloat{appfig}{h}{loappfig}[chapter]}
\floatname{appfig}{Figure W}
\newcommand*\quartoappfigref[1]{Figure \hyperref[#1]{W\ref{#1}}}
\@ifpackageloaded{caption}{}{\usepackage{caption}}
\DeclareCaptionLabelFormat{quartoappfigreflabelformat}{#1#2}
\makeatother
\makeatletter
\@ifpackageloaded{float}{}{\usepackage{float}}
\floatstyle{plain}
\@ifundefined{c@chapter}{\newfloat{appenv}{h}{loapp}}{\newfloat{appenv}{h}{loapp}[chapter]}
\floatname{appenv}{Web Appendix}
\floatstyle{plaintop}
\restylefloat{appenv}
\newcommand*\quartoappref[1]{Web \hyperref[#1]{Appendix\ref{#1}}}
\@ifpackageloaded{caption}{}{\usepackage{caption}}
\DeclareCaptionLabelFormat{quartoappreflabelformat}{#1#2}
\makeatother
\makeatletter
\@ifpackageloaded{caption}{}{\usepackage{caption}}
\AtBeginDocument{%
\ifdefined\contentsname
  \renewcommand*\contentsname{Table of contents}
\else
  \newcommand\contentsname{Table of contents}
\fi
\ifdefined\listfigurename
  \renewcommand*\listfigurename{List of Figures}
\else
  \newcommand\listfigurename{List of Figures}
\fi
\ifdefined\listtablename
  \renewcommand*\listtablename{List of Tables}
\else
  \newcommand\listtablename{List of Tables}
\fi
\ifdefined\figurename
  \renewcommand*\figurename{\textbf{Fig.}}
\else
  \newcommand\figurename{\textbf{Fig.}}
\fi
\ifdefined\tablename
  \renewcommand*\tablename{\textbf{Table}}
\else
  \newcommand\tablename{\textbf{Table}}
\fi
}
\@ifpackageloaded{float}{}{\usepackage{float}}
\floatstyle{ruled}
\@ifundefined{c@chapter}{\newfloat{codelisting}{h}{lop}}{\newfloat{codelisting}{h}{lop}[chapter]}
\floatname{codelisting}{Listing}

\makeatother
\makeatletter
\usepackage{pdflscape}
\makeatother
\makeatletter
\@ifpackageloaded{caption}{}{\usepackage{caption}}
\@ifpackageloaded{subcaption}{}{\usepackage{subcaption}}
\makeatother
\usepackage{bookmark}
\IfFileExists{xurl.sty}{\usepackage{xurl}}{} % add URL line breaks if available
\makeatletter
\@ifundefined{xmpquote}{}{}
\makeatother
\hypersetup{
  pdftitle={Does Rank Still Matter? Position Bias When AI Agents Shop on Our Behalf},
  pdfauthor={Davood Wadi; Yu Ma},
  colorlinks=true,
  linkcolor={blue},
  filecolor={Maroon},
  citecolor={Blue},
  urlcolor={Blue},
  pdfcreator={LaTeX via pandoc}}

\title{Does Rank Still Matter? Position Bias When AI Agents Shop on Our
Behalf}
\author{Davood Wadi \and Yu Ma}
\date{}
\begin{document}
\maketitle

\textbf{Abstract}

When shopping is delegated to AI agents, it is unclear whether the
ranking advantage documented for humans persists. Across 7,000 sessions
with five large language models (LLMs) and varying reasoning effort, we
compare AI agents with human field data. AI agents search more
extensively than human consumers. As with humans, lower-ranked listings
are less likely to be inspected, although the effect is smaller for AI
agents. Unlike humans, AI agents show a pattern consistent with the
lost-in-the-middle effect, whereby middle listings have the lowest
probability of inspection. At the choice stage, position effects are
concentrated at lower reasoning effort, but higher effort reduces the
middle penalty for every LLM that exhibits it. These findings suggest
that, when search is delegated, displayed attributes may matter more
than placement, and that exposure to position bias depends on how the AI
agent is configured.

\textbf{Keywords:} search rankings, AI agents, large language models,
search advertising, consideration sets, delegated consumption

\section{Introduction}\label{introduction}

Search rankings are a strategic marketing variable. Search engine
optimization budgets and the premium paid for top slots rely on the
well-documented finding that people examine and purchase higher-placed
alternatives more often
\citep{narayanan2015position, joachims2007evaluating, donnelly2024welfare}.
The canonical explanation is that ranking determines which alternatives
enter the consideration set while leaving the consumer's valuation of an
inspected alternative unchanged
\citep{hauser2014consideration, akchen2026consider, ursu2018power}. This
is because scanning a listing page is sequential, scrolling takes effort
\citep{joachims2007evaluating, greminger2026trade}, multi-attribute
comparison is cognitively demanding \citep{punj2009information}, and
attention is depleted over the course of a search episode
\citep{ursu2023search}. As a result, human consumers tend to terminate
their search after examining very few alternatives. For example, in
hotel search, the average impression yields only 1.12 clicks
\citep{ursu2018power}. Rankings are thus valuable precisely because
human attention is scarce, sequential, and easily exhausted.

Consumers are now delegating parts of this search to AI agents
\citep{balaskas2026recommendations, wadi2026shopping}. Industry
analytics show that 38\% of U.S. consumers have used generative AI for
online shopping, most often for product research and recommendations,
and that referral traffic from generative AI platforms to U.S. retail
sites grew by an order of magnitude between mid-2024 and mid-2025
\citep{adobe2025genai}. Rather than sifting a results page themselves,
consumers increasingly rely on conversational agents that retrieve,
compare, and transact on their behalf
\citep{mogaji2024generative, lee2026semantic, kumar2026transformative, hasselwander2026toward}.

Unlike a human consumer, an AI agent receives the entire results page at
once within its context window \citep{hengle2026can}, so inspecting a
listing at the bottom of the page requires no scrolling. If the value of
rank stems from limited human attention, delegation to AI agents should,
in principle, eliminate it. However, the large language models (LLMs)
that power the AI agents display heuristics and biases of their own
\citep{brady2025dual, wadi2026every, macmillan2024ir, wadi2026shopping, allouah2026your},
including a tendency to retrieve information placed in the middle of a
long context less reliably than information at either end
\citep{liu2024lost, su2024roformer}. Rank may therefore continue to
shape delegated search through a mechanism that is unrelated to
scrolling.

Moreover, because human consumers typically click on a single
alternative \citep{honka2024consumer}, an effect of rank on clicking is
effectively an effect on choice. When an AI agent inspects several
alternatives, however, this correspondence may no longer hold, and it
remains unclear whether a positional advantage at the inspection stage
carries over to the final choice. Finally, unlike human attention, the
deliberation of an LLM is configurable. Users can set the reasoning
effort that the model expends before acting
\citep{xu2025toward, alomrani2025reasoning}, which raises the question
of whether position effects can be mitigated through configuration of
the LLM.

We therefore address three research questions:

\begin{itemize}
\tightlist
\item
  \textbf{RQ1.} Does presentation order shape what an AI agent inspects,
  as it does for human consumers?
\item
  \textbf{RQ2.} Does position affect the final choice?
\item
  \textbf{RQ3.} Does increasing the reasoning effort change the position
  effect on choice?
\end{itemize}

To examine these questions, we place LLM agents in a simulated hotel
search environment designed to be comparable to human field data
\citep{ursu2018power}. In each session, an AI agent receives a results
page of 100 hotel listings presented in a randomized order, inspects
hotels through a tool call analogous to a click, and submits a booking
through a separate tool call. The main experiment comprises 7,000
sessions across five proprietary LLMs from Google, Anthropic, and
OpenAI, each tested at two to four levels of reasoning effort.
Additional experiments examine whether the findings are robust to the
wording of the prompt, the framing of the booking tool, and the display
of hotel names (1,500 sessions). We extend the analysis to four
open-weight LLMs (2,000 sessions).

The findings show that AI agents search more extensively than human
consumers, inspecting between 1.63 and 5.83 listings per session at
their default reasoning effort (compared with 1.12 for human consumers).
Furthermore, whereas human consumers decline to book in a third of
sessions, the AI agents booked a hotel in most sessions. This pattern
persisted when the instruction to recommend a hotel was removed and the
booking tool explicitly allowed the AI agent to end the session without
booking.

Position continues to shape which listings are inspected, although this
position penalty is less than half of that observed for human consumers.
However, rather than declining monotonically, the probability of
inspection reaches its minimum between ranks 62 and 81 and rises toward
the bottom of the page---a pattern consistent with the
lost-in-the-middle effect documented for LLMs \citep{liu2024lost}. At
the choice stage, the influence of position varies across LLMs but is
concentrated at lower levels of reasoning effort. For the LLMs that
display the middle penalty, increasing the reasoning effort
significantly reduces the middle penalty on choice.

These findings contribute to research on consumption search in three
ways. First, they distinguish the value of rank from its conventional
explanation. Position effects among human consumers are attributed to
the costs of scrolling, which are absent for an AI agent that receives
the entire listing page at once, yet position continues to shape
inspection. Unlike humans, the positional disadvantage is strongest in
the middle of the page, which is consistent with the retrieval
limitations of LLMs. Second, they show that the close correspondence
between consideration and choice observed in human search does not
extend to AI-delegated search. Because AI agents inspect several
alternatives, position can shape the composition of their consideration
sets without determining their final choice. Third, they identify
reasoning effort as a strong moderator of position effects. Because
reasoning effort is configurable, the extent to which delegated choices
depend on position is determined not just by which LLM is deployed but
also by how it is configured.

The findings also carry implications for sellers and for the platforms
and consumers who delegate search. For sellers, the middle rather than
the bottom of the results page is the least advantageous placement when
search is delegated to AI agents. The advantage of the top rank depends
on the LLM and its configuration, which suggests that search engine
optimization strategies developed for human consumers require
recalibration. For platforms and consumers, the results qualify the
common practice of deploying a more capable LLM to reduce bias, because
position-neutral choice depends on reasoning effort rather than LLM
capability. As a result, position-neutral choices can be obtained at
approximately 4\% of the cost of the most capable LLM tested.

\section{Experiment}\label{experiment}

\subsection{Design}\label{design}

To simulate the sequential search process, the hotel information was
partitioned into two layers. The initial prompt displayed the hotel
listing page (e.g., \emph{hotel name}, \emph{review score}, \emph{price
nightly}, and \emph{price total} for each hotel). The AI agent
interacted with the environment through tool calls. A tool is a named
function whose purpose and arguments are described to the LLM. Instead
of replying in text, the LLM can request that the function be executed,
whereupon the environment executes the request and returns the result to
the LLM, which continues its reasoning with this information. To obtain
more information about a hotel, the LLM was required to utilize the
\emph{inspect} tool that resembled human clicks. Upon calling the
\emph{inspect} tool with a specific hotel's ID, the environment returned
detailed information about that hotel. This included the hotel's star
rating, specific amenities, granular sub-ratings for cleanliness and
service, and detailed room availability and room price. To book a hotel,
the LLM called a separate tool (\emph{submit\_choice}) with the ID of
the chosen hotel, which recorded the booking and ended the session. To
eliminate the endogeneity bias inherent in search engine ranking
algorithms, where higher-quality options are systematically placed at
the top, the presentation order of the 100 hotels was fully randomized
for each session.

We prompted the AI agent with a persona of a hotel booking assistant
delegated by a traveler to evaluate options and book a room based on the
trip parameters (see Table~\ref{tbl-prompt-comparison}, ``Main''). Trip
parameters were chosen to match the modal search in the human benchmark
\citep{ursu2018power}: an itinerary for two adults, no children, one
room, and a two-night stay including a Saturday night (check-in
2026-05-30, check-out 2026-06-01).

\begin{longtable}[]{@{}
  >{\raggedright\arraybackslash}p{(\linewidth - 2\tabcolsep) * \real{0.1250}}
  >{\raggedright\arraybackslash}p{(\linewidth - 2\tabcolsep) * \real{0.8750}}@{}}
\caption{Comparison of prompt
instructions}\label{tbl-prompt-comparison}\tabularnewline
\toprule\noalign{}
\begin{minipage}[b]{\linewidth}\raggedright
Condition
\end{minipage} & \begin{minipage}[b]{\linewidth}\raggedright
System instructions
\end{minipage} \\
\midrule\noalign{}
\endfirsthead
\toprule\noalign{}
\begin{minipage}[b]{\linewidth}\raggedright
Condition
\end{minipage} & \begin{minipage}[b]{\linewidth}\raggedright
System instructions
\end{minipage} \\
\midrule\noalign{}
\endhead
\bottomrule\noalign{}
\endlastfoot
Main & \emph{You have been delegated by a traveler to book a hotel in
Manhattan, New York. Trip: 2 adults, 1 room, check-in 2026-05-30,
check-out 2026-06-01 (Saturday night stay). A list of hotels from
Expedia is shown below. You can see each hotel's description from the
search results page. To view the full details of a hotel (amenities,
location, sub-ratings, policies), call the inspect tool with the hotel's
option ID. Once you have enough information, call submit\_choice with
your final decision. Which hotel would you book for the traveler?} \\
Alternative & \emph{You have been delegated by a traveler to book a
hotel in Manhattan, New York. Trip: 2 adults, 1 room, check-in
2026-05-30, check-out 2026-06-01 (Saturday night stay). A list of hotels
from Expedia is shown below. You can call the inspect tool with the
hotel's option ID to view the full details of a hotel (amenities,
location, sub-ratings, policies), or you can call submit\_choice with
your final decision.} \\
Finish & \emph{You have been delegated by a traveler planning a stay in
Manhattan, New York. Trip: 2 adults, 1 room, check-in 2026-05-30,
check-out 2026-06-01 (Saturday night stay). A list of hotels from
Expedia is shown below. You can call the inspect tool with the hotel's
option ID to view the full details of a hotel (amenities, location,
sub-ratings, policies), or you can call finish with your final
decision.} \\
\end{longtable}

\emph{Note.} The ``Main'' condition includes an instruction to stop once
the agent has enough information and closes with an explicit
recommendation question (i.e., asking which hotel to book for the
traveler). The ``Alternative'' condition removes the stopping rule and
the recommendation question. The ``Finish'' condition removes the
booking instruction and replaces \emph{submit\_choice} with a
\emph{finish} tool whose hotel argument is optional, so the agent can
end the session without booking.

\subsection{Sample and procedure}\label{sample-and-procedure}

Because LLM providers are phasing out user control over sampling
parameters such as temperature and top-p
\citep{google2026sampling, anthropic2026sampling}, the configurations
that remain under the control of a consumer or platform are the choice
of LLM and its reasoning effort. Therefore, sampling parameters were
left at the providers' defaults (Web Appendix D), and we manipulated the
LLM and its reasoning effort. Our sample consisted of proprietary LLMs
from Google (Gemini 3.1 Flash Lite, Gemini 3.7 Flash, and Gemini 3.1
Pro), Anthropic (Claude Sonnet 5), and OpenAI (GPT 6 Luna). Providers
expose different reasoning effort levels, so these levels are comparable
within, but not across, LLMs. We tested minimal (default), low, medium,
and high for Flash Lite; low and medium (default) for Flash; low,
medium, and high (default) for Pro; low and high (default) for Sonnet;
and none, medium (default), and xHigh for Luna.

Following Monte Carlo approaches that use repeated sampling to reliably
assess LLM behavior \citep{wadi2025monte}, we collected 500 independent
sessions for each of the 14 experimental cells (\emph{LLM} ×
\emph{effort}; 7,000 sessions total). In each session, the
\emph{inspect} tool could be called any number of times, and a booking
could be submitted at any point with the \emph{submit\_choice} tool.
Ending the session without a booking represented the outside option. The
order of all tool calls was recorded.

\subsection{Measures}\label{measures}

For each session, every displayed hotel contributed one observation
indicating whether it was inspected and whether it was ultimately
chosen. \emph{Inspected} was coded 1 when the AI agent invoked the
\emph{inspect} tool for a given hotel and 0 otherwise (an analogue of a
click in the human benchmark). \emph{Chosen} was coded 1 when the hotel
was chosen and 0 otherwise. Because the 100 hotel observations within a
session were not independent, standard errors in all models were
clustered at the session level.

The main independent variable was position, indexed from 1 to 100 based
on the hotel's randomized placement on the listing page shown to the AI
agent. We also used the attributes that were directly observable on the
listing page as control variables. \emph{Price} captured the nightly
price for the hotel, while \emph{Review score} captured the aggregate
guest rating shown on the search-results page. We added two additional
listing-page controls to approximate the human study's covariates.
\emph{Chain} was coded as a binary indicator for the presence of a major
hotel brand name in the listing title. \emph{Promotion} was coded as a
binary indicator for the presence of a visible discount badge on the
listing page \citep[for details see Web Appendix C]{ursu2018power}.

\subsection{Results}\label{results}

At their provider-default reasoning effort, AI agents search
considerably more than human consumers
(Table~\ref{tbl-descriptive-stats};
\quartoappfigref{appfig-insp-hist})\footnote{The human benchmark
  displays approximately 27 listings per results page, includes only
  impressions with at least one click, and controls for query
  characteristics and destination, which our design holds fixed. The two
  environments are compared in \quartoapptblref{apptbl-env-attributes}.}.
Whereas 93\% of human sessions contained a single inspection, such
sessions accounted for only 0.2\% to 37.4\% of LLM sessions. Moreover,
none of the LLMs selected the outside option in any condition, whereas
human consumers who clicked at least once did not book a hotel in 34\%
of sessions.

\begin{longtable}[]{@{}
  >{\raggedright\arraybackslash}p{(\linewidth - 12\tabcolsep) * \real{0.3462}}
  >{\raggedright\arraybackslash}p{(\linewidth - 12\tabcolsep) * \real{0.0962}}
  >{\raggedright\arraybackslash}p{(\linewidth - 12\tabcolsep) * \real{0.0769}}
  >{\raggedright\arraybackslash}p{(\linewidth - 12\tabcolsep) * \real{0.1346}}
  >{\raggedright\arraybackslash}p{(\linewidth - 12\tabcolsep) * \real{0.0865}}
  >{\raggedright\arraybackslash}p{(\linewidth - 12\tabcolsep) * \real{0.0769}}
  >{\raggedright\arraybackslash}p{(\linewidth - 12\tabcolsep) * \real{0.1827}}@{}}
\caption{Search behavior for LLMs (default reasoning) versus human
consumers}\label{tbl-descriptive-stats}\tabularnewline
\toprule\noalign{}
\begin{minipage}[b]{\linewidth}\raggedright
\end{minipage} & \begin{minipage}[b]{\linewidth}\raggedright
Sonnet
\end{minipage} & \begin{minipage}[b]{\linewidth}\raggedright
Luna
\end{minipage} & \begin{minipage}[b]{\linewidth}\raggedright
Flash Lite
\end{minipage} & \begin{minipage}[b]{\linewidth}\raggedright
Flash
\end{minipage} & \begin{minipage}[b]{\linewidth}\raggedright
Pro
\end{minipage} & \begin{minipage}[b]{\linewidth}\raggedright
Human Benchmark
\end{minipage} \\
\midrule\noalign{}
\endfirsthead
\toprule\noalign{}
\begin{minipage}[b]{\linewidth}\raggedright
\end{minipage} & \begin{minipage}[b]{\linewidth}\raggedright
Sonnet
\end{minipage} & \begin{minipage}[b]{\linewidth}\raggedright
Luna
\end{minipage} & \begin{minipage}[b]{\linewidth}\raggedright
Flash Lite
\end{minipage} & \begin{minipage}[b]{\linewidth}\raggedright
Flash
\end{minipage} & \begin{minipage}[b]{\linewidth}\raggedright
Pro
\end{minipage} & \begin{minipage}[b]{\linewidth}\raggedright
Human Benchmark
\end{minipage} \\
\midrule\noalign{}
\endhead
\bottomrule\noalign{}
\endlastfoot
\emph{N} & 500 & 500 & 500 & 500 & 500 & 166,036 \\
Conversion rate & 100.0\% & 100.0\% & 100.0\% & 100.0\% & 100.0\% &
66.0\% \\
Outside option selected & 0.0\% & 0.0\% & 0.0\% & 0.0\% & 0.0\% &
34.0\% \\
Inspections per session & & & & & & \\
(\emph{Mean}) & 1.63 & 4.00 & 3.12 & 4.25 & 5.83 & 1.12 \\
(\emph{Median}) & 2.00 & 4.00 & 3.00 & 4.00 & 5.00 & 1.00 \\
(\emph{Mode}) & 2.00 & 4.00 & 3.00 & 4.00 & 1.00 & 1.00 \\
(\emph{SD}) & 0.49 & 1.13 & 0.54 & 1.59 & 4.78 & 0.61 \\
Sessions with one inspection & 37.4\% & 0.2\% & 0.8\% & 2.6\% & 20.8\% &
93.0\% \\
Chose the first inspected hotel & 98.2\% & 15.6\% & 62.0\% & 99.8\% &
60.4\% & --- \\
Position of the chosen hotel (\emph{M}) & 47.58 & 33.45 & 43.97 & 49.46
& 49.72 & --- \\
Modal choice proportion & 95.8\% & 41.0\% & 61.4\% & 100.0\% & 55.4\% &
--- \\
\end{longtable}

\emph{Note.} LLMs are shown at their provider-default reasoning effort;
statistics for all effort levels are reported in
\quartoapptblref{apptbl-effort-descriptive}. Human benchmark data
adapted from \citet{ursu2018power}, Table 1; the human sample includes
only impressions with one or more clicks. Em-dashes (---) indicate
metrics not available in the human field data. The distribution of Pro
is right-skewed (\quartoappfigref{appfig-insp-hist}).

\subsubsection{Position effect}\label{position-effect}

We estimate the effect of position with linear probability models
(ordinary least squares regressions of a binary outcome) of
\emph{Inspected} and \emph{Chosen}, separately for each LLM × effort
condition, controlling for price, review score, chain, and promotion.
The coefficient on position in the linear specification measures the
position penalty---the reduction in the probability of inspection or
choice from moving a listing down by one rank---which is directly
comparable with the human benchmark. Because the position penalty cannot
distinguish a monotone decline from a U-shaped relationship, we also
estimate a linear probability model that adds the square of position
(rescaled to the unit interval) to measure the middle penalty. This
quadratic term measures the extent to which listings in the middle of
the page fall below the straight line between the top and the bottom, or
the lost-in-the-middle effect \citep{liu2024lost}.\footnote{With
  position \(x\) scaled to the unit interval, the predicted probability
  is \(f(x) = \beta_1 x + \beta_2 x^2\) plus the remaining terms of the
  model. The decline from the top to the bottom of the page is
  \(f(0) - f(1) = -(\beta_1 + \beta_2)\), which equals 100 times the
  position penalty because position is randomized uniformly. The middle
  penalty---the shortfall of the midpoint relative to the straight line
  between the top and the bottom---is
  \([f(0) + f(1)]/2 - f(0.5) = \beta_2 / 4\).}

\begin{figure}

\centering{

\pandocbounded{\includegraphics[keepaspectratio]{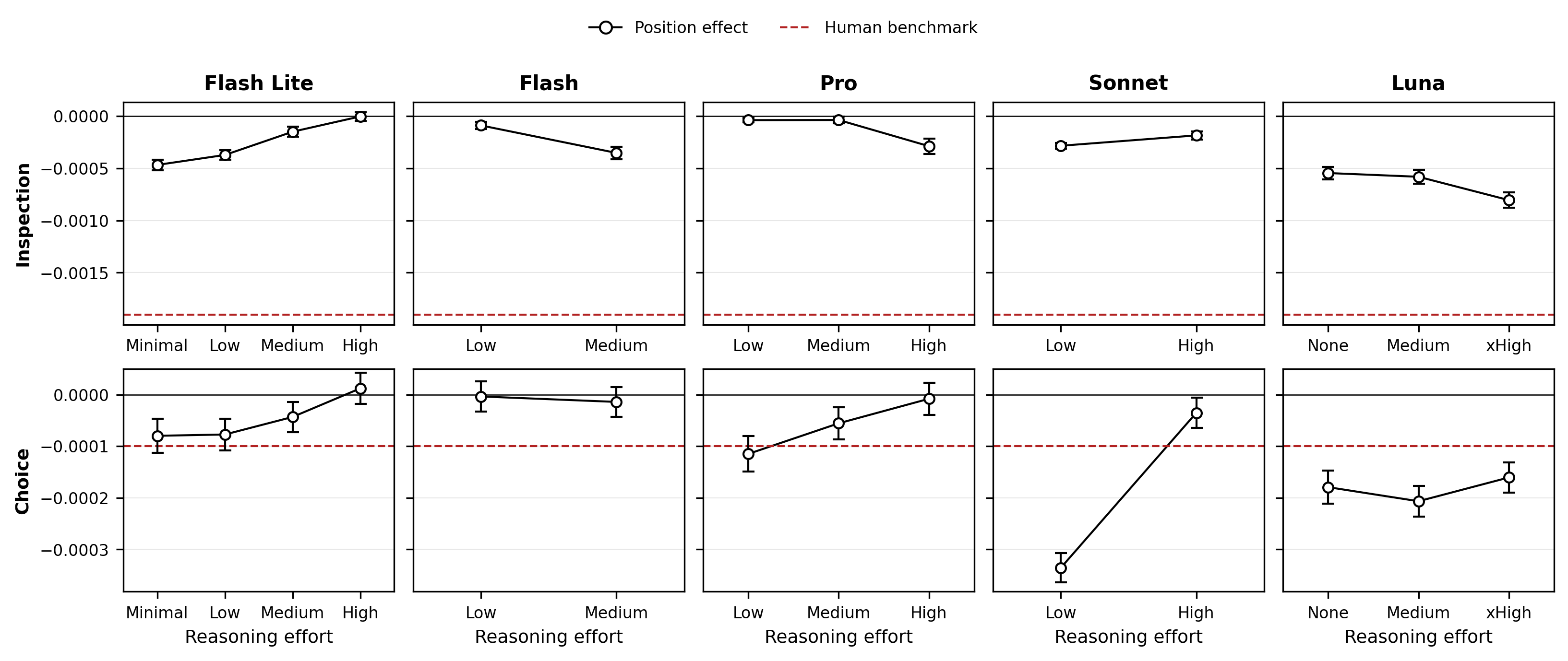}}

}

\caption{\label{fig-position-effect}Position effect on inspection and
choice by reasoning effort, compared with human consumers}

\end{figure}%

\emph{Note.} Points are coefficients on position from linear probability
models estimated separately for each LLM and reasoning effort level
(\quartoapptblref{apptbl-ols-inspected};
\quartoapptblref{apptbl-ols-chosen}), expressed as the change in the
probability of inspection or choice from moving a listing down by one
rank. For inspection, a value of −0.0005 indicates that each rank lower
reduces the probability that a listing is inspected by 0.05 percentage
points. For choice, a value of −0.0001 indicates a reduction of 0.01
percentage points per rank. Models control for price, review score,
chain, and promotion. The dashed line marks the human benchmark from
\citet{ursu2018power}, Table 2. Error bars are 95\% confidence intervals
based on standard errors clustered at the session level.

\begin{figure}

\centering{

\pandocbounded{\includegraphics[keepaspectratio]{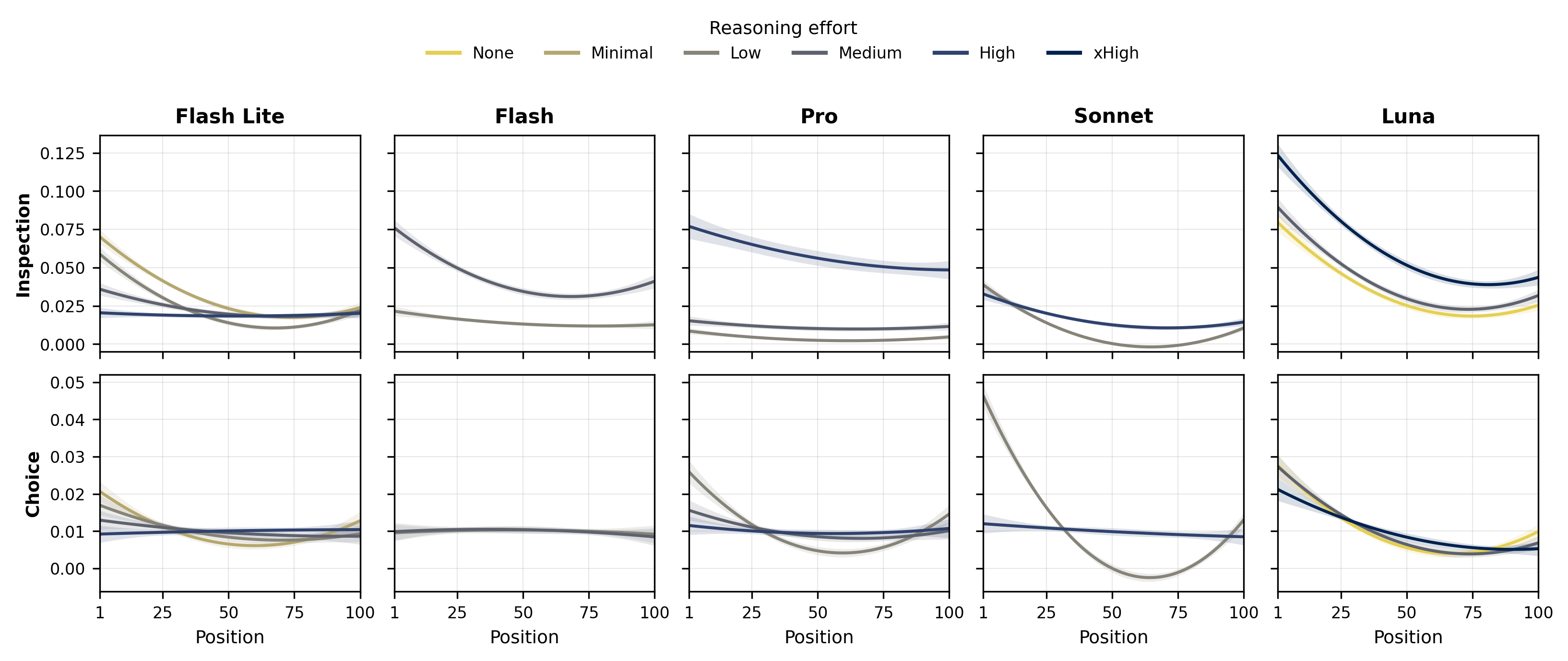}}

}

\caption{\label{fig-effort-emm}Estimated marginal means of the quadratic
position effect under varying reasoning effort levels}

\end{figure}%

\emph{Note.} Lines are predicted probabilities from linear probability
models with position and position\(^2\)
(\quartoapptblref{apptbl-effort-ols-sq-inspected};
\quartoapptblref{apptbl-effort-ols-sq-chosen}), with controls held at
their sample means. Shaded bands are 95\% confidence intervals based on
standard errors clustered at the session level.

For inspection, the position penalty is small relative to that of human
consumers. Moving a listing lower by one rank reduces its probability of
inspection by at most 0.0008 for any LLM at any effort level, compared
with 0.0019 for human consumers (Fig.~\ref{fig-position-effect};
\quartoapptblref{apptbl-ols-inspected}). For humans, the probability of
clicking declines monotonically with rank, which makes the bottom of the
listing page the least advantageous placement
(\quartoapptblref{apptbl-ols-inspected}). In contrast, for AI agents,
the probability of inspection reaches its minimum between ranks 62 and
81 and rises toward the bottom (Fig.~\ref{fig-effort-emm}), with a
significant middle penalty in 12 of the 14 LLM × effort conditions (RQ1;
\quartoapptblref{apptbl-effort-ols-sq-inspected}). Reasoning effort
affects the middle penalty in inspection without a consistent
direction.\footnote{To test whether reasoning effort alters the middle
  penalty, we pool all effort levels of each LLM into a single linear
  probability model in which the position terms are interacted with
  effort. Effort is treated as a categorical variable with the lowest
  level as the reference. A joint Wald test then assesses whether these
  interactions differ from zero.}

For choice, the position penalty ranges from statistically
indistinguishable from zero to approximately three times that of human
consumers (Fig.~\ref{fig-position-effect};
\quartoapptblref{apptbl-ols-chosen}). The middle penalty in choice is
significant in 8 of the 14 conditions, predominantly at lower levels of
reasoning effort, with the probability of choice reaching its minimum
between ranks 60 and 90 (RQ2; Fig.~\ref{fig-effort-emm};
\quartoapptblref{apptbl-effort-ols-sq-chosen}). Importantly, higher
reasoning effort reduces the middle penalty for every LLM that exhibits
the penalty (Flash Lite, \(\chi^2\)(6) = 51.11; Pro, \(\chi^2\)(4) =
54.85; Sonnet, \(\chi^2\)(2) = 292.33; all \emph{p}s \textless{} 0.001;
Luna, \(\chi^2\)(4) = 17.70, \emph{p} \textless{} 0.01). The middle
penalty becomes non-significant at the highest effort level for Flash
Lite, Pro, and Sonnet and is reduced by 63\% for Luna. Flash exhibits no
middle penalty in choice at either effort level (RQ3).

\subsubsection{Cost of position-neutral
choice}\label{cost-of-position-neutral-choice}

For sellers and consumers, the relevant question is whether a delegated
AI shopping agent's choice depends on position at all. We define choice
as position-neutral when the probability that a listing is chosen does
not depend on its position on the listing page, conditional on its
displayed attributes (price, review score, chain, and promotion). We
assess this with a joint test that the coefficients on position and
position\(^2\) in the quadratic linear probability model of choice are
both zero (\quartoapptblref{apptbl-effort-cost}). We treat a
configuration as position-neutral when this test does not reject the
null hypothesis at the 5\% level, and position-dependent otherwise.

Among these configurations, Flash Lite at high effort is the least
expensive position-neutral configuration, at \$0.011 per session, which
amounts to 4\% of the cost of Pro (\$0.28) and 12\% of the cost of
Sonnet (\$0.096) at high effort, whereas Pro at low effort remains
strongly position-dependent (\(\chi^2\)(2) = 107.21, \emph{p}
\textless{} 0.001; \quartoapptblref{apptbl-effort-cost}). Raising the
effort of Flash Lite from minimal to high does not increase its cost
(\$0.013 versus \$0.011 per session), because the additional reasoning
tokens are offset by fewer inspections (3.12 versus 1.91 per session).
Luna, in contrast, remains position-dependent at every effort level (all
\emph{p}s \textless{} 0.001).

\subsubsection{Robustness}\label{robustness}

We next examine whether the wording of the prompt, the framing of the
booking tool, and the display of hotel names could account for the main
findings. Because replicating each check across all 14 LLM × effort
conditions would require 21,000 additional sessions, we conduct each
check at the default reasoning effort of the LLM that is most
susceptible to the corresponding confound, where an artifact of the
design would be most readily detected.

A first concern is that two instructions in the main prompt (i.e.,
asking the agent which hotel it would book and instructing it to stop
once it has sufficient information) may have shaped the AI agents'
behavior. The former instruction could inflate the conversion rate,
while the latter could imply that several hotels should be examined,
thereby increasing search relative to human consumers. We conducted this
check with Flash Lite, whose significant curvilinear position effect on
both inspection and choice at its default reasoning effort would make
any artifact of the prompt most apparent.

We executed 500 sessions with a reduced prompt that states the trip
parameters and the two available tools but omits both instructions
(Table~\ref{tbl-prompt-comparison}, Alternative). Under this prompt, the
conversion rate remains at 100\%, and search remains deeper than in the
human benchmark (\quartoapptblref{apptbl-robustness-descriptive}).
Moreover, the curvilinear position effect replicates for both inspection
and choice, with a negative linear term and a positive quadratic term
(all \emph{p}s \textless{} 0.001;
\quartoapptblref{apptbl-robustness-ols}).

A second concern is that, even in the absence of an explicit
recommendation question, a tool named \emph{submit\_choice} may frame
the termination of a session as the act of choosing, thereby
discouraging the outside option. Therefore, we tested the \emph{Finish}
condition, in which we removed the booking instruction and replaced
\emph{submit\_choice} with a \emph{finish} tool whose hotel argument is
optional, which allows the AI agent to end the session without booking
(Table~\ref{tbl-prompt-comparison}, Finish). We conducted this test with
Flash because its choices are concentrated on a single hotel under the
\emph{Main} condition. Thus, any shift toward the outside option would
be attributable to the framing of the booking tool.

Across 500 sessions, all sessions concluded with a booking
(\quartoapptblref{apptbl-robustness-descriptive}). The curvilinear
effect of position on inspection persisted, with a negative linear and a
positive quadratic term (both \emph{p}s \textless{} 0.001). Similar to
the \emph{Main} condition, choice showed no detectable dependence on
position (\(\chi^2\)(2) = 2.71, \emph{p} = 0.258;
\quartoapptblref{apptbl-robustness-ols}). These results indicate that
the absence of the outside option is not an artifact of the booking
tool.

A third concern is that the concentration of Flash's choices on a single
listing may reflect familiarity with the hotel's brand, acquired during
pretraining, rather than the hotel attributes displayed on the listing
page. In the \emph{No name} condition, we therefore removed hotel names
from the listing page while retaining every other attribute and the main
prompt.

The percentage of the modal hotel choice was 93.8\%, compared with 100\%
under the \emph{Main} condition
(\quartoapptblref{apptbl-robustness-descriptive}). As in the main
condition, position reduced the probability of inspection but had no
detectable effect on choice (\quartoapptblref{apptbl-robustness-ols}).
This modest decline suggests that hotel names contribute partly to the
choice but that the convergence on a single listing is driven
predominantly by its displayed attributes.

\subsubsection{Open-weight models}\label{open-weight-models}

To assess whether the findings extend beyond proprietary LLMs, we
replicated the \emph{Main} condition with four open-weight LLMs: Meta's
Muse Glimmer 30B, NVIDIA's Nemotron 3.5 Lightning 30B A3B, Alibaba
Cloud's Qwen3.8 27B, and Google's Gemma 4 31B (500 sessions each;
\quartoapptblref{apptbl-open-descriptive}).

Search remains deeper than in the human benchmark, although the outside
option is selected only by Nemotron, in 2.2\% of sessions. For
inspection, the linear term on position is negative and significant for
every LLM (all \emph{p}s \textless{} 0.01), and the quadratic term is
positive and significant for three of the four LLMs
(\quartoapptblref{apptbl-open-ols}). At the choice stage, Muse Glimmer
exhibits the curvilinear pattern observed for Flash Lite and Luna,
whereas neither position term is significant for Gemma and Nemotron.
Qwen shows a small positive linear term (\emph{p} \textless{} 0.05).

\section{Conclusion}\label{conclusion}

By randomizing the order of the hotel listings presented to AI agents,
this research studied the value of rank for AI shopping agents. We found
that, similar to humans, AI shopping agents can display a position bias
(RQ1 and RQ2), but the nature and ramifications of this bias are
different. AI agents' position bias is concentrated in the middle rather
than at the bottom of the page \citep{ursu2018power, liu2024lost}.
Moreover, unlike human consideration sets, which typically hold a single
alternative, AI agents inspected several listings. As a result, position
could shape the AI agents' consideration sets without affecting their
final choices, particularly at higher levels of reasoning effort (RQ2).

For platforms and consumers, these findings suggest that
position-neutral choice can be obtained more economically by raising the
reasoning effort of a less capable LLM than by deploying a more capable
one. For example, Pro, the most capable Google LLM tested, remained
strongly position-dependent at low effort at a cost of \$0.037 per
session, whereas Flash Lite at high effort showed no detectable
dependence on position at \$0.011 per session. Similarly, Pro attained
position-neutral choice only at high effort, at a cost of \$0.28 per
session, approximately 25 times that of Flash Lite at high effort.
However, the mitigating effect of reasoning effort was not uniform
across LLMs, as Luna remained position-dependent even at its highest
effort level. Therefore, platforms and consumers should verify the
susceptibility to position effects of the specific LLM and configuration
they deploy.

These findings carry implications beyond position effects. AI agents
concentrated their bookings on listings with high review scores and low
prices (\quartoapptblref{apptbl-ols-chosen}), which mirrors human
consumers \citep{chevalier2006effect, floyd2014online} and is consistent
with other studies of AI shopping agents
\citep{allouah2026your, cherep2026framework, wadi2026whom, kireyev2026pricebench}.
Sellers may therefore benefit more from strengthening these signals
\citep[e.g., by soliciting
reviews;][]{godes2009firm, wadi2026interplay, wadi2026careful, burtch2018stimulating}
than from improving their placement.

Our study has several limitations. We examined a single booking task
with a fixed query and destination. Categories that differ in the number
or complexity of attributes may produce different position effects.
Moreover, we observe a pattern consistent with the lost-in-the-middle
phenomenon but do not identify retrieval as the mechanism, which would
require access to the internal activations of the LLMs. Costs are based
on list prices, which change over time and do not reflect prompt
caching, and the LLMs tested are moving targets whose positional biases
may change with future releases. Future research could examine sponsored
and personalized rankings, multi-page results, and the strategies
sellers adopt when they optimize listings for AI agents rather than for
human consumers.

\section{Data availability statement}\label{data-availability-statement}

The data and the code to reproduce the analysis are available at the
following anonymous repository:

https://osf.io/vg8sn/overview?view\_only=93557b5ec07745ceb19bf61e88f65ee1

\section{References}\label{references}

\renewcommand{\bibsection}{}
\bibliography{refs/ai-marketing.bib,refs/acl-refs.bib,refs/marketing.bib,refs/others.bib,refs/myrefs.bib,refs/seo.bib,refs/mouselab.bib}

\begin{thebibliography}{38}
\providecommand{\natexlab}[1]{#1}
\providecommand{\url}[1]{\texttt{#1}}
\expandafter\ifx\csname urlstyle\endcsname\relax
  \providecommand{\doi}[1]{doi: #1}\else
  \providecommand{\doi}{doi: \begingroup \urlstyle{rm}\Url}\fi

\bibitem[Akchen and Mitrofanov(2026)]{akchen2026consider}
Yi-Chun Akchen and Dmitry Mitrofanov.
\newblock Consider or choose? the role and power of consideration sets.
\newblock \emph{Management Science}, 72\penalty0 (7):\penalty0 6025--6042,
  2026.

\bibitem[Allouah et~al.(2026)Allouah, Besbes, Figueroa, Kanoria, and
  Kumar]{allouah2026your}
Amine Allouah, Omar Besbes, Josu{\'e}~D Figueroa, Yash Kanoria, and Akshit
  Kumar.
\newblock What is your ai agent buying? evaluation, biases, model dependence,
  \& emerging implications of agentic e-commerce.
\newblock In \emph{Proceedings of the ACM Web Conference 2026}, pages
  8697--8700, 2026.

\bibitem[Alomrani et~al.(2025)Alomrani, Zhang, Li, Sun, Pal, Zhang, Hu, Ajwani,
  Valkanas, Karimi, et~al.]{alomrani2025reasoning}
Mohammad~Ali Alomrani, Yingxue Zhang, Derek Li, Qianyi Sun, Soumyasundar Pal,
  Zhanguang Zhang, Yaochen Hu, Rohan~Deepak Ajwani, Antonios Valkanas, Raika
  Karimi, et~al.
\newblock Reasoning on a budget: A survey of adaptive and controllable
  test-time compute in llms.
\newblock \emph{arXiv preprint arXiv:2507.02076}, 2025.

\bibitem[{Anthropic}(2026)]{anthropic2026sampling}
{Anthropic}.
\newblock Model deprecations: {API} parameter deprecations, 2026.
\newblock URL
  \url{https://platform.claude.com/docs/en/about-claude/model-deprecations#api-parameter-deprecations}.

\bibitem[Balaskas(2026)]{balaskas2026recommendations}
Stefanos Balaskas.
\newblock From recommendations to delegation: A systematic review mapping
  agentic ai in e-commerce and its consumer effects.
\newblock \emph{Inf.}, 17\penalty0 (3):\penalty0 222, 2026.

\bibitem[Brady et~al.(2025)Brady, Nulty, Zhang, Ward, and
  McGovern]{brady2025dual}
Oliver Brady, Paul Nulty, Lili Zhang, Tom{\'a}s~E Ward, and David~P McGovern.
\newblock Dual-process theory and decision-making in large language models.
\newblock \emph{Nature Reviews Psychology}, 4\penalty0 (12):\penalty0 777--792,
  2025.

\bibitem[Burtch et~al.(2018)Burtch, Hong, Bapna, and
  Griskevicius]{burtch2018stimulating}
Gordon Burtch, Yili Hong, Ravi Bapna, and Vladas Griskevicius.
\newblock Stimulating online reviews by combining financial incentives and
  social norms.
\newblock \emph{Management Science}, 64\penalty0 (5):\penalty0 2065--2082,
  2018.

\bibitem[Cherep et~al.(2026)Cherep, Ma, Xu, Shaked, Maes, and
  Singh]{cherep2026framework}
Manuel Cherep, Chengtian Ma, Abigail Xu, Maya Shaked, Pattie Maes, and Nikhil
  Singh.
\newblock A framework for studying ai agent behavior: Evidence from consumer
  choice experiments.
\newblock In \emph{International Conference on Learning Representations},
  volume 2026, pages 63296--63326, 2026.

\bibitem[Chevalier and Mayzlin(2006)]{chevalier2006effect}
Judith~A Chevalier and Dina Mayzlin.
\newblock The effect of word of mouth on sales: Online book reviews.
\newblock \emph{Journal of marketing research}, 43\penalty0 (3):\penalty0
  345--354, 2006.

\bibitem[Donnelly et~al.(2024)Donnelly, Kanodia, and
  Morozov]{donnelly2024welfare}
Robert Donnelly, Ayush Kanodia, and Ilya Morozov.
\newblock Welfare effects of personalized rankings.
\newblock \emph{Marketing Science}, 43\penalty0 (1):\penalty0 92--113, 2024.

\bibitem[Floyd et~al.(2014)Floyd, Freling, Alhoqail, Cho, and
  Freling]{floyd2014online}
Kristopher Floyd, Ryan Freling, Saad Alhoqail, Hyun~Young Cho, and Traci
  Freling.
\newblock How online product reviews affect retail sales: A meta-analysis.
\newblock \emph{Journal of retailing}, 90\penalty0 (2):\penalty0 217--232,
  2014.

\bibitem[Godes and Mayzlin(2009)]{godes2009firm}
David Godes and Dina Mayzlin.
\newblock Firm-created word-of-mouth communication: Evidence from a field test.
\newblock \emph{Marketing science}, 28\penalty0 (4):\penalty0 721--739, 2009.

\bibitem[{Google}(2026)]{google2026sampling}
{Google}.
\newblock Gemini models: Sampling parameter deprecation, 2026.
\newblock URL
  \url{https://ai.google.dev/gemini-api/docs/latest-model#sampling-parameter-deprecation}.

\bibitem[Greminger(2026)]{greminger2026trade}
Rafael~P Greminger.
\newblock Trade-offs between ranking objectives: Descriptive evidence and
  structural estimation.
\newblock \emph{Management Science}, 2026.

\bibitem[Hasselwander et~al.(2026)Hasselwander, Sunio, Lah, and
  Mogaji]{hasselwander2026toward}
Marc Hasselwander, Varsolo Sunio, Oliver Lah, and Emmanuel Mogaji.
\newblock Toward agentic ai: User acceptance of a deeply personalized ai super
  assistant (aisa).
\newblock \emph{Journal of Retailing and Consumer Services}, 89:\penalty0
  104620, 2026.

\bibitem[Hauser(2014)]{hauser2014consideration}
John~R Hauser.
\newblock Consideration-set heuristics.
\newblock \emph{Journal of Business Research}, 67\penalty0 (8):\penalty0
  1688--1699, 2014.

\bibitem[Hengle et~al.(2026)Hengle, Bajpai, Dan, and
  Chakraborty]{hengle2026can}
Amey Hengle, Prasoon Bajpai, Soham Dan, and Tanmoy Chakraborty.
\newblock Can llms reason over extended multilingual contexts? towards
  long-context evaluation beyond retrieval over haystacks.
\newblock In \emph{Proceedings of the 19th Conference of the European Chapter
  of the Association for Computational Linguistics (Volume 1: Long Papers)},
  pages 6128--6152, 2026.

\bibitem[Honka et~al.(2024)Honka, Seiler, and Ursu]{honka2024consumer}
Elisabeth Honka, Stephan Seiler, and Raluca Ursu.
\newblock Consumer search: What can we learn from pre-purchase data?
\newblock \emph{Journal of Retailing}, 100\penalty0 (1):\penalty0 114--129,
  2024.

\bibitem[Joachims et~al.(2007)Joachims, Granka, Pan, Hembrooke, Radlinski, and
  Gay]{joachims2007evaluating}
Thorsten Joachims, Laura Granka, Bing Pan, Helene Hembrooke, Filip Radlinski,
  and Geri Gay.
\newblock Evaluating the accuracy of implicit feedback from clicks and query
  reformulations in web search.
\newblock \emph{ACM Transactions on Information Systems (TOIS)}, 25\penalty0
  (2):\penalty0 7--es, 2007.

\bibitem[Kireyev(2026)]{kireyev2026pricebench}
Pavel Kireyev.
\newblock Pricebench: A diagnostic benchmark for price, quality, and brand
  preferences in llm booking agents.
\newblock In \emph{Forthcoming, Proceedings of the 2026 Conference on Empirical
  Methods in Natural Language Processing: Industry Track (EMNLP 2026)}, 2026.

\bibitem[Kumar et~al.(2026)Kumar, Kotler, and Kumar]{kumar2026transformative}
V~Kumar, Philip Kotler, and Ajay Kumar.
\newblock Transformative marketing strategies in the era of new-age
  technologies: Principles, plan, purpose, and practice.
\newblock \emph{Journal of the Academy of Marketing Science}, 54\penalty0
  (1):\penalty0 1--27, 2026.

\bibitem[Lee et~al.(2026)Lee, Lee, and Suh]{lee2026semantic}
Woo-Chul Lee, Jang-Suk Lee, and Jungho Suh.
\newblock Semantic divergence in ai-generated and human influencer product
  recommendations: A computational analysis of dual-agent communication in
  social commerce.
\newblock \emph{Applied Sciences}, 16\penalty0 (6):\penalty0 2816, 2026.

\bibitem[Liu et~al.(2024)Liu, Lin, Hewitt, Paranjape, Bevilacqua, Petroni, and
  Liang]{liu2024lost}
Nelson~F Liu, Kevin Lin, John Hewitt, Ashwin Paranjape, Michele Bevilacqua,
  Fabio Petroni, and Percy Liang.
\newblock Lost in the middle: How language models use long contexts.
\newblock \emph{Transactions of the association for computational linguistics},
  12:\penalty0 157--173, 2024.

\bibitem[Macmillan-Scott and Musolesi(2024)]{macmillan2024ir}
Olivia Macmillan-Scott and Mirco Musolesi.
\newblock (ir) rationality and cognitive biases in large language models.
\newblock \emph{Royal Society open science}, 11\penalty0 (6), 2024.

\bibitem[Mogaji and Jain(2024)]{mogaji2024generative}
Emmanuel Mogaji and Varsha Jain.
\newblock How generative ai is (will) change consumer behaviour: Postulating
  the potential impact and implications for research, practice, and policy.
\newblock \emph{Journal of consumer behaviour}, 23\penalty0 (5):\penalty0
  2379--2389, 2024.

\bibitem[Narayanan and Kalyanam(2015)]{narayanan2015position}
Sridhar Narayanan and Kirthi Kalyanam.
\newblock Position effects in search advertising and their moderators: A
  regression discontinuity approach.
\newblock \emph{Marketing Science}, 34\penalty0 (3):\penalty0 388--407, 2015.

\bibitem[Pandya()]{adobe2025genai}
Vivek Pandya.
\newblock Generative {AI}-powered shopping rises with traffic to {U.S.} retail
  sites up 4,700\%.
\newblock URL
  \url{https://business.adobe.com/blog/generative-ai-powered-shopping-rises-with-traffic-to-retail-sites}.

\bibitem[Punj and Moore(2009)]{punj2009information}
Girish Punj and Robert Moore.
\newblock Information search and consideration set formation in a web-based
  store environment.
\newblock \emph{Journal of Business Research}, 62\penalty0 (6):\penalty0
  644--650, 2009.

\bibitem[Su et~al.(2024)Su, Ahmed, Lu, Pan, Bo, and Liu]{su2024roformer}
Jianlin Su, Murtadha Ahmed, Yu~Lu, Shengfeng Pan, Wen Bo, and Yunfeng Liu.
\newblock Roformer: Enhanced transformer with rotary position embedding.
\newblock \emph{Neurocomputing}, 568:\penalty0 127063, 2024.

\bibitem[Ursu(2018)]{ursu2018power}
Raluca~M Ursu.
\newblock The power of rankings: Quantifying the effect of rankings on online
  consumer search and purchase decisions.
\newblock \emph{Marketing Science}, 37\penalty0 (4):\penalty0 530--552, 2018.

\bibitem[Ursu et~al.(2023)Ursu, Zhang, and Honka]{ursu2023search}
Raluca~M Ursu, Qianyun Zhang, and Elisabeth Honka.
\newblock Search gaps and consumer fatigue.
\newblock \emph{Marketing Science}, 42\penalty0 (1):\penalty0 110--136, 2023.

\bibitem[Wadi and Fredette(2025)]{wadi2025monte}
Davood Wadi and Marc Fredette.
\newblock A monte-carlo sampling framework for reliable evaluation of large
  language models using behavioral analysis.
\newblock \emph{Findings of the Association for Computational Linguistics:
  EMNLP 2025}, pages 9414--9432, 2025.

\bibitem[Wadi and Ma(2026{\natexlab{a}})]{wadi2026shopping}
Davood Wadi and Yu~Ma.
\newblock Shopping by algorithm: How agentic ai deploys human heuristics as a
  surrogate consumer.
\newblock 2026{\natexlab{a}}.

\bibitem[Wadi and Ma(2026{\natexlab{b}})]{wadi2026whom}
Davood Wadi and Yu~Ma.
\newblock Whom do ai agents work for? role assignment induces sponsorship bias
  in llm recommenders.
\newblock \emph{arXiv preprint arXiv:2609.17989}, 2026{\natexlab{b}}.

\bibitem[Wadi et~al.(2026{\natexlab{a}})Wadi, Fredette, Senecal, and
  Legoux]{wadi2026careful}
Davood Wadi, Marc Fredette, Sylvain Senecal, and Renaud Legoux.
\newblock Be careful what you pay for: the effect of performance contingent
  incentives on online product reviews.
\newblock \emph{Journal of Research in Interactive Marketing}, pages 1--25,
  2026{\natexlab{a}}.

\bibitem[Wadi et~al.(2026{\natexlab{b}})Wadi, Ghodrat, and
  Philp]{wadi2026every}
Davood Wadi, Mohsen Ghodrat, and Matthew Philp.
\newblock Every token counts: Exact likert-scale distributions for measuring
  llm attitudes and biases.
\newblock \emph{arXiv preprint arXiv:2608.10503}, 2026{\natexlab{b}}.

\bibitem[Wadi et~al.(2026{\natexlab{c}})Wadi, Legoux, Fredette, and
  S{\'e}n{\'e}cal]{wadi2026interplay}
Davood Wadi, Renaud Legoux, Marc Fredette, and Sylvain S{\'e}n{\'e}cal.
\newblock The interplay of altruism and financial incentives: Maximizing online
  reviews through effective messaging.
\newblock \emph{Journal of Electronic Commerce Research}, 27\penalty0 (2),
  2026{\natexlab{c}}.

\bibitem[Xu et~al.(2025)Xu, Hao, Shao, Zong, Li, Wang, Zhang, Wang, Lan, Gong,
  et~al.]{xu2025toward}
Fengli Xu, Qianyue Hao, Chenyang Shao, Zefang Zong, Yu~Li, Jingwei Wang, Yunke
  Zhang, Jingyi Wang, Xiaochong Lan, Jiahui Gong, et~al.
\newblock Toward large reasoning models: A survey of reinforced reasoning with
  large language models.
\newblock \emph{Patterns}, 6\penalty0 (10), 2025.

\end{thebibliography}

\newpage{}

\section*{Web Appendix A - Supplementary tables}\label{sec-appendix-a}
\addcontentsline{toc}{section}{Web Appendix A - Supplementary tables}

{\def\LTcaptype{apptbl}
\begin{longtable}[]{@{}
  >{\raggedright\arraybackslash}p{(\linewidth - 4\tabcolsep) * \real{0.3125}}
  >{\raggedright\arraybackslash}p{(\linewidth - 4\tabcolsep) * \real{0.3250}}
  >{\raggedright\arraybackslash}p{(\linewidth - 4\tabcolsep) * \real{0.3625}}@{}}
\caption{{\label{apptbl-env-attributes}}Comparison of search
environment hotel attributes between current study and human benchmark}\tabularnewline
\toprule\noalign{}
\begin{minipage}[b]{\linewidth}\raggedright
Hotel listing attribute
\end{minipage} & \begin{minipage}[b]{\linewidth}\raggedright
Current study
\end{minipage} & \begin{minipage}[b]{\linewidth}\raggedright
Human benchmark
\end{minipage} \\
\midrule\noalign{}
\endfirsthead
\toprule\noalign{}
\begin{minipage}[b]{\linewidth}\raggedright
Hotel listing attribute
\end{minipage} & \begin{minipage}[b]{\linewidth}\raggedright
Current study
\end{minipage} & \begin{minipage}[b]{\linewidth}\raggedright
Human benchmark
\end{minipage} \\
\midrule\noalign{}
\endhead
\bottomrule\noalign{}
\endlastfoot
Price per night (\$) & \$230.64 (96.86) & \$159.71 (102.43) \\
Review score (1--5) & 4.32 (0.29) & 3.89 (0.86) \\
Chain proxy & 55.0\% & 66.0\% \\
Promotion proxy & 19.0\% & 25.0\% \\
\end{longtable}
}

\emph{Note.} Values represent sample means with standard deviations in
parentheses where applicable. Human benchmark data adapted from
\citet{ursu2018power}, Table 1.

\begin{landscape}

{\def\LTcaptype{apptbl}
\begin{longtable}[]{@{}
  >{\raggedright\arraybackslash}p{(\linewidth - 18\tabcolsep) * \real{0.0915}}
  >{\raggedright\arraybackslash}p{(\linewidth - 18\tabcolsep) * \real{0.0915}}
  >{\raggedright\arraybackslash}p{(\linewidth - 18\tabcolsep) * \real{0.0493}}
  >{\raggedright\arraybackslash}p{(\linewidth - 18\tabcolsep) * \real{0.0775}}
  >{\raggedright\arraybackslash}p{(\linewidth - 18\tabcolsep) * \real{0.0634}}
  >{\raggedright\arraybackslash}p{(\linewidth - 18\tabcolsep) * \real{0.0493}}
  >{\raggedright\arraybackslash}p{(\linewidth - 18\tabcolsep) * \real{0.1197}}
  >{\raggedright\arraybackslash}p{(\linewidth - 18\tabcolsep) * \real{0.1690}}
  >{\raggedright\arraybackslash}p{(\linewidth - 18\tabcolsep) * \real{0.1268}}
  >{\raggedright\arraybackslash}p{(\linewidth - 18\tabcolsep) * \real{0.1056}}@{}}
\caption{{\label{apptbl-effort-descriptive}}Descriptive statistics by
reasoning effort}\tabularnewline
\toprule\noalign{}
\multirow{2}{=}{\begin{minipage}[b]{\linewidth}\raggedright
LLM
\end{minipage}} &
\multirow{2}{=}{\begin{minipage}[b]{\linewidth}\raggedright
Effort
\end{minipage}} &
\multicolumn{4}{>{\raggedright\arraybackslash}p{(\linewidth - 18\tabcolsep) * \real{0.2394} + 6\tabcolsep}}{%
\begin{minipage}[b]{\linewidth}\raggedright
Inspections per session
\end{minipage}} & \begin{minipage}[b]{\linewidth}\raggedright
One inspection
\end{minipage} & \begin{minipage}[b]{\linewidth}\raggedright
Chose first inspected
\end{minipage} & \begin{minipage}[b]{\linewidth}\raggedright
Chosen position
\end{minipage} & \begin{minipage}[b]{\linewidth}\raggedright
Modal choice
\end{minipage} \\
& & \begin{minipage}[b]{\linewidth}\raggedright
\emph{Mean}
\end{minipage} & \begin{minipage}[b]{\linewidth}\raggedright
\emph{Median}
\end{minipage} & \begin{minipage}[b]{\linewidth}\raggedright
\emph{Mode}
\end{minipage} & \begin{minipage}[b]{\linewidth}\raggedright
\emph{SD}
\end{minipage} & \begin{minipage}[b]{\linewidth}\raggedright
\%
\end{minipage} & \begin{minipage}[b]{\linewidth}\raggedright
\%
\end{minipage} & \begin{minipage}[b]{\linewidth}\raggedright
\emph{M}
\end{minipage} & \begin{minipage}[b]{\linewidth}\raggedright
\%
\end{minipage} \\
\midrule\noalign{}
\endfirsthead
\toprule\noalign{}
\multirow{2}{=}{\begin{minipage}[b]{\linewidth}\raggedright
LLM
\end{minipage}} &
\multirow{2}{=}{\begin{minipage}[b]{\linewidth}\raggedright
Effort
\end{minipage}} &
\multicolumn{4}{>{\raggedright\arraybackslash}p{(\linewidth - 18\tabcolsep) * \real{0.2394} + 6\tabcolsep}}{%
\begin{minipage}[b]{\linewidth}\raggedright
Inspections per session
\end{minipage}} & \begin{minipage}[b]{\linewidth}\raggedright
One inspection
\end{minipage} & \begin{minipage}[b]{\linewidth}\raggedright
Chose first inspected
\end{minipage} & \begin{minipage}[b]{\linewidth}\raggedright
Chosen position
\end{minipage} & \begin{minipage}[b]{\linewidth}\raggedright
Modal choice
\end{minipage} \\
& & \begin{minipage}[b]{\linewidth}\raggedright
\emph{Mean}
\end{minipage} & \begin{minipage}[b]{\linewidth}\raggedright
\emph{Median}
\end{minipage} & \begin{minipage}[b]{\linewidth}\raggedright
\emph{Mode}
\end{minipage} & \begin{minipage}[b]{\linewidth}\raggedright
\emph{SD}
\end{minipage} & \begin{minipage}[b]{\linewidth}\raggedright
\%
\end{minipage} & \begin{minipage}[b]{\linewidth}\raggedright
\%
\end{minipage} & \begin{minipage}[b]{\linewidth}\raggedright
\emph{M}
\end{minipage} & \begin{minipage}[b]{\linewidth}\raggedright
\%
\end{minipage} \\
\midrule\noalign{}
\endhead
\bottomrule\noalign{}
\endlastfoot
Flash Lite & Minimal\textsuperscript{†} & 3.12 & 3.00 & 3.00 & 0.54 &
0.8\% & 62.0\% & 43.97 & 61.4\% \\
& Low & 2.28 & 2.00 & 2.00 & 1.05 & 24.0\% & 61.4\% & 44.02 & 78.8\% \\
& Medium & 2.26 & 2.00 & 2.00 & 1.10 & 27.0\% & 85.4\% & 46.32 &
83.8\% \\
& High & 1.91 & 2.00 & 2.00 & 0.66 & 22.8\% & 96.6\% & 51.20 & 96.6\% \\
Flash & Low & 1.45 & 1.00 & 1.00 & 0.71 & 65.8\% & 99.8\% & 50.44 &
100.0\% \\
& Medium\textsuperscript{†} & 4.25 & 4.00 & 4.00 & 1.59 & 2.6\% & 99.8\%
& 49.46 & 100.0\% \\
Pro & Low & 0.39 & 0.00 & 0.00 & 0.55 & 35.8\% & 36.0\% & 40.96 &
48.0\% \\
& Medium & 1.13 & 1.00 & 1.00 & 1.28 & 53.6\% & 62.0\% & 45.63 &
51.8\% \\
& High\textsuperscript{†} & 5.83 & 5.00 & 1.00 & 4.78 & 20.8\% & 60.4\%
& 49.72 & 55.4\% \\
Sonnet & Low & 0.85 & 1.00 & 1.00 & 0.48 & 75.6\% & 76.6\% & 22.63 &
30.6\% \\
& High\textsuperscript{†} & 1.63 & 2.00 & 2.00 & 0.49 & 37.4\% & 98.2\%
& 47.58 & 95.8\% \\
Luna & None & 3.43 & 3.00 & 3.00 & 1.01 & 5.2\% & 28.8\% & 35.56 &
27.6\% \\
& Medium\textsuperscript{†} & 4.00 & 4.00 & 4.00 & 1.13 & 0.2\% & 15.6\%
& 33.45 & 41.0\% \\
& xHigh & 6.22 & 6.00 & 6.00 & 1.89 & 0.0\% & 13.0\% & 36.99 & 47.4\% \\
\end{longtable}
}

\emph{Note.} Every condition has 500 sessions, a 100.0\% conversion
rate, and 0.0\% outside option selection. \textsuperscript{†} Provider
default reasoning effort. One inspection = share of sessions with a
single inspection; Chose first inspected = share of sessions booking the
first hotel inspected; Modal choice = share of sessions booking the most
frequently chosen hotel.

{\def\LTcaptype{apptbl}
\begin{longtable}[]{@{}
  >{\raggedright\arraybackslash}p{(\linewidth - 16\tabcolsep) * \real{0.0922}}
  >{\raggedright\arraybackslash}p{(\linewidth - 16\tabcolsep) * \real{0.0922}}
  >{\raggedright\arraybackslash}p{(\linewidth - 16\tabcolsep) * \real{0.1277}}
  >{\raggedright\arraybackslash}p{(\linewidth - 16\tabcolsep) * \real{0.1277}}
  >{\raggedright\arraybackslash}p{(\linewidth - 16\tabcolsep) * \real{0.1064}}
  >{\raggedright\arraybackslash}p{(\linewidth - 16\tabcolsep) * \real{0.1135}}
  >{\raggedright\arraybackslash}p{(\linewidth - 16\tabcolsep) * \real{0.1135}}
  >{\raggedright\arraybackslash}p{(\linewidth - 16\tabcolsep) * \real{0.0922}}
  >{\raggedright\arraybackslash}p{(\linewidth - 16\tabcolsep) * \real{0.0851}}@{}}
\caption{{\label{apptbl-ols-inspected}}Linear probability models of
inspection by LLM and reasoning effort, with the human benchmark}\tabularnewline
\toprule\noalign{}
\begin{minipage}[b]{\linewidth}\raggedright
LLM
\end{minipage} & \begin{minipage}[b]{\linewidth}\raggedright
Effort
\end{minipage} & \begin{minipage}[b]{\linewidth}\raggedright
Position
\end{minipage} & \begin{minipage}[b]{\linewidth}\raggedright
Price
\end{minipage} & \begin{minipage}[b]{\linewidth}\raggedright
Review score
\end{minipage} & \begin{minipage}[b]{\linewidth}\raggedright
Chain
\end{minipage} & \begin{minipage}[b]{\linewidth}\raggedright
Promotion
\end{minipage} & \begin{minipage}[b]{\linewidth}\raggedright
Adj. \(R^2\)
\end{minipage} & \begin{minipage}[b]{\linewidth}\raggedright
\emph{N}
\end{minipage} \\
\midrule\noalign{}
\endfirsthead
\toprule\noalign{}
\begin{minipage}[b]{\linewidth}\raggedright
LLM
\end{minipage} & \begin{minipage}[b]{\linewidth}\raggedright
Effort
\end{minipage} & \begin{minipage}[b]{\linewidth}\raggedright
Position
\end{minipage} & \begin{minipage}[b]{\linewidth}\raggedright
Price
\end{minipage} & \begin{minipage}[b]{\linewidth}\raggedright
Review score
\end{minipage} & \begin{minipage}[b]{\linewidth}\raggedright
Chain
\end{minipage} & \begin{minipage}[b]{\linewidth}\raggedright
Promotion
\end{minipage} & \begin{minipage}[b]{\linewidth}\raggedright
Adj. \(R^2\)
\end{minipage} & \begin{minipage}[b]{\linewidth}\raggedright
\emph{N}
\end{minipage} \\
\midrule\noalign{}
\endhead
\bottomrule\noalign{}
\endlastfoot
Flash Lite & Minimal\textsuperscript{†} & -0.0005*** (\textless0.0001) &
-0.0001*** (\textless0.0001) & 0.1466*** (0.0014) & -0.0645*** (0.0013)
& -0.0042** (0.0014) & 0.0891 & 50,000 \\
& Low & -0.0004*** (\textless0.0001) & -0.0001*** (\textless0.0001) &
0.1025*** (0.0021) & -0.0391*** (0.0013) & 0.0146*** (0.0014) & 0.0621 &
50,000 \\
& Medium & -0.0001*** (\textless0.0001) & -0.0001*** (\textless0.0001) &
0.1026*** (0.0023) & -0.0387*** (0.0014) & 0.0162*** (0.0013) & 0.0591 &
50,000 \\
& High & \textgreater-0.0001 (\textless0.0001) & -0.0001***
(\textless0.0001) & 0.0854*** (0.0013) & -0.0322*** (0.0010) & 0.0211***
(0.0008) & 0.0523 & 50,000 \\
Flash & Low & -0.0001*** (\textless0.0001) & \textgreater-0.0001***
(\textless0.0001) & 0.0654*** (0.0015) & -0.0225*** (0.0009) & 0.0351***
(0.0008) & 0.0547 & 50,000 \\
& Medium\textsuperscript{†} & -0.0004*** (\textless0.0001) & -0.0001***
(\textless0.0001) & 0.1967*** (0.0033) & -0.0839*** (0.0020) &
-0.0186*** (0.0018) & 0.1077 & 50,000 \\
Pro & Low & \textgreater-0.0001*** (\textless0.0001) &
\textgreater-0.0001*** (\textless0.0001) & 0.0123*** (0.0009) &
-0.0054*** (0.0006) & 0.0053*** (0.0010) & 0.0075 & 50,000 \\
& Medium & \textgreater-0.0001* (\textless0.0001) & -0.0001***
(\textless0.0001) & 0.0301*** (0.0016) & -0.0152*** (0.0010) & 0.0095***
(0.0015) & 0.0154 & 50,000 \\
& High\textsuperscript{†} & -0.0003*** (\textless0.0001) & -0.0003***
(\textless0.0001) & 0.1282*** (0.0058) & -0.0387*** (0.0023) & -0.0067**
(0.0025) & 0.0344 & 50,000 \\
Sonnet & Low & -0.0003*** (\textless0.0001) & \textgreater-0.0001***
(\textless0.0001) & 0.0369*** (0.0011) & -0.0146*** (0.0008) & -0.0006
(0.0014) & 0.0254 & 50,000 \\
& High\textsuperscript{†} & -0.0002*** (\textless0.0001) &
\textgreater-0.0001*** (\textless0.0001) & 0.0757*** (0.0012) &
-0.0304*** (0.0008) & 0.0265*** (0.0009) & 0.0573 & 50,000 \\
Luna & None & -0.0005*** (\textless0.0001) & -0.0001***
(\textless0.0001) & 0.1079*** (0.0016) & -0.0035* (0.0014) & 0.0220***
(0.0013) & 0.0399 & 50,000 \\
& Medium\textsuperscript{†} & -0.0006*** (\textless0.0001) & -0.0001***
(\textless0.0001) & 0.1328*** (0.0020) & -0.0061*** (0.0015) & 0.0111***
(0.0014) & 0.0456 & 50,000 \\
& xHigh & -0.0008*** (\textless0.0001) & -0.0002*** (\textless0.0001) &
0.2037*** (0.0030) & -0.0041* (0.0019) & -0.0100*** (0.0018) & 0.0659 &
50,000 \\
Human & & -0.0019*** (\textless0.0001) & -0.0001*** (\textless0.0001) &
0.0012*** (0.0002) & 0.0022*** (0.0005) & 0.0116*** (0.0005) & 0.0150 &
1,220,917 \\
\end{longtable}
}

\emph{Note.} Coefficients are from linear probability models. Standard
errors clustered at the session level in parentheses.
\textsuperscript{†} Provider default reasoning effort. Human benchmark
from \citet{ursu2018power}, Table 2. Price is reported in \$1 units to
match the human benchmark. Controls comprise hotel attributes displayed
on the listing page; star rating and location score, which the human
benchmark's results page displayed, are not shown on Expedia's current
results page. Query characteristics and destination fixed effects are
omitted because our design holds the query and destination fixed.\\
*p\textless0.05, **p\textless0.01, ***p\textless0.001

{\def\LTcaptype{apptbl}
\begin{longtable}[]{@{}
  >{\raggedright\arraybackslash}p{(\linewidth - 16\tabcolsep) * \real{0.0935}}
  >{\raggedright\arraybackslash}p{(\linewidth - 16\tabcolsep) * \real{0.0935}}
  >{\raggedright\arraybackslash}p{(\linewidth - 16\tabcolsep) * \real{0.1151}}
  >{\raggedright\arraybackslash}p{(\linewidth - 16\tabcolsep) * \real{0.1295}}
  >{\raggedright\arraybackslash}p{(\linewidth - 16\tabcolsep) * \real{0.1079}}
  >{\raggedright\arraybackslash}p{(\linewidth - 16\tabcolsep) * \real{0.1151}}
  >{\raggedright\arraybackslash}p{(\linewidth - 16\tabcolsep) * \real{0.1151}}
  >{\raggedright\arraybackslash}p{(\linewidth - 16\tabcolsep) * \real{0.0935}}
  >{\raggedright\arraybackslash}p{(\linewidth - 16\tabcolsep) * \real{0.0863}}@{}}
\caption{{\label{apptbl-ols-chosen}}Linear probability models of choice
by LLM and reasoning effort, with the human benchmark}\tabularnewline
\toprule\noalign{}
\begin{minipage}[b]{\linewidth}\raggedright
LLM
\end{minipage} & \begin{minipage}[b]{\linewidth}\raggedright
Effort
\end{minipage} & \begin{minipage}[b]{\linewidth}\raggedright
Position
\end{minipage} & \begin{minipage}[b]{\linewidth}\raggedright
Price
\end{minipage} & \begin{minipage}[b]{\linewidth}\raggedright
Review score
\end{minipage} & \begin{minipage}[b]{\linewidth}\raggedright
Chain
\end{minipage} & \begin{minipage}[b]{\linewidth}\raggedright
Promotion
\end{minipage} & \begin{minipage}[b]{\linewidth}\raggedright
Adj. \(R^2\)
\end{minipage} & \begin{minipage}[b]{\linewidth}\raggedright
\emph{N}
\end{minipage} \\
\midrule\noalign{}
\endfirsthead
\toprule\noalign{}
\begin{minipage}[b]{\linewidth}\raggedright
LLM
\end{minipage} & \begin{minipage}[b]{\linewidth}\raggedright
Effort
\end{minipage} & \begin{minipage}[b]{\linewidth}\raggedright
Position
\end{minipage} & \begin{minipage}[b]{\linewidth}\raggedright
Price
\end{minipage} & \begin{minipage}[b]{\linewidth}\raggedright
Review score
\end{minipage} & \begin{minipage}[b]{\linewidth}\raggedright
Chain
\end{minipage} & \begin{minipage}[b]{\linewidth}\raggedright
Promotion
\end{minipage} & \begin{minipage}[b]{\linewidth}\raggedright
Adj. \(R^2\)
\end{minipage} & \begin{minipage}[b]{\linewidth}\raggedright
\emph{N}
\end{minipage} \\
\midrule\noalign{}
\endhead
\bottomrule\noalign{}
\endlastfoot
Flash Lite & Minimal\textsuperscript{†} & -0.0001*** (\textless0.0001) &
\textgreater-0.0001*** (\textless0.0001) & 0.0470*** (0.0003) &
-0.0184*** (0.0005) & 0.0169*** (0.0016) & 0.0351 & 50,000 \\
& Low & -0.0001*** (\textless0.0001) & \textgreater-0.0001***
(\textless0.0001) & 0.0449*** (0.0002) & -0.0146*** (0.0004) & 0.0300***
(0.0013) & 0.0431 & 50,000 \\
& Medium & \textgreater-0.0001** (\textless0.0001) &
\textgreater-0.0001*** (\textless0.0001) & 0.0448*** (0.0002) &
-0.0146*** (0.0004) & 0.0331*** (0.0012) & 0.0467 & 50,000 \\
& High & \textless0.0001 (\textless0.0001) & \textgreater-0.0001***
(\textless0.0001) & 0.0442*** (0.0001) & -0.0128*** (0.0002) & 0.0421***
(0.0006) & 0.0570 & 50,000 \\
Flash & Low & \textgreater-0.0001 (\textless0.0001) &
\textgreater-0.0001*** (\textless0.0001) & 0.0439*** (\textless0.0001) &
-0.0124*** (\textless0.0001) & 0.0445*** (\textless0.0001) & 0.0602 &
50,000 \\
& Medium\textsuperscript{†} & \textgreater-0.0001 (\textless0.0001) &
\textgreater-0.0001*** (\textless0.0001) & 0.0439*** (\textless0.0001) &
-0.0124*** (\textless0.0001) & 0.0445*** (\textless0.0001) & 0.0602 &
50,000 \\
Pro & Low & -0.0001*** (\textless0.0001) & \textgreater-0.0001***
(\textless0.0001) & 0.0322*** (0.0010) & -0.0124*** (0.0009) & 0.0177***
(0.0016) & 0.0217 & 50,000 \\
& Medium & -0.0001*** (\textless0.0001) & \textgreater-0.0001***
(\textless0.0001) & 0.0324*** (0.0009) & -0.0139*** (0.0009) & 0.0175***
(0.0017) & 0.0224 & 50,000 \\
& High\textsuperscript{†} & \textgreater-0.0001 (\textless0.0001) &
\textgreater-0.0001*** (\textless0.0001) & 0.0264*** (0.0012) &
-0.0031** (0.0010) & 0.0311*** (0.0015) & 0.0242 & 50,000 \\
Sonnet & Low & -0.0003*** (\textless0.0001) & \textgreater-0.0001***
(\textless0.0001) & 0.0422*** (0.0006) & -0.0151*** (0.0008) & 0.0025
(0.0015) & 0.0289 & 50,000 \\
& High\textsuperscript{†} & \textgreater-0.0001* (\textless0.0001) &
\textgreater-0.0001*** (\textless0.0001) & 0.0439*** (0.0001) &
-0.0128*** (0.0002) & 0.0420*** (0.0006) & 0.0567 & 50,000 \\
Luna & None & -0.0002*** (\textless0.0001) & \textgreater-0.0001***
(\textless0.0001) & 0.0319*** (0.0008) & 0.0008 (0.0009) & 0.0066***
(0.0013) & 0.0121 & 50,000 \\
& Medium\textsuperscript{†} & -0.0002*** (\textless0.0001) &
\textgreater-0.0001*** (\textless0.0001) & 0.0274*** (0.0006) &
0.0071*** (0.0008) & -0.0060*** (0.0008) & 0.0121 & 50,000 \\
& xHigh & -0.0002*** (\textless0.0001) & \textgreater-0.0001***
(\textless0.0001) & 0.0261*** (0.0004) & 0.0093*** (0.0006) & -0.0023**
(0.0008) & 0.0106 & 50,000 \\
Human & & -0.0001*** (\textless0.0001) & \textgreater-0.0001***
(\textless0.0001) & 0.0002*** (\textless0.0001) & 0.0003* (0.0001) &
0.0012*** (0.0001) & 0.0030 & 1,220,917 \\
\end{longtable}
}

\emph{Note.} Coefficients are from linear probability models. Standard
errors clustered at the session level in parentheses.
\textsuperscript{†} Provider default reasoning effort. Human benchmark
from \citet{ursu2018power}, Table 2. Price is reported in \$1 units to
match the human benchmark. Controls comprise hotel attributes displayed
on the listing page; star rating and location score, which the human
benchmark's results page displayed, are not shown on Expedia's current
results page. Query characteristics and destination fixed effects are
omitted because our design holds the query and destination fixed.\\
*p\textless0.05, **p\textless0.01, ***p\textless0.001

{\def\LTcaptype{apptbl}
\begin{longtable}[]{@{}
  >{\raggedright\arraybackslash}p{(\linewidth - 16\tabcolsep) * \real{0.0915}}
  >{\raggedright\arraybackslash}p{(\linewidth - 16\tabcolsep) * \real{0.0915}}
  >{\raggedright\arraybackslash}p{(\linewidth - 16\tabcolsep) * \real{0.1127}}
  >{\raggedright\arraybackslash}p{(\linewidth - 16\tabcolsep) * \real{0.1056}}
  >{\raggedright\arraybackslash}p{(\linewidth - 16\tabcolsep) * \real{0.1268}}
  >{\raggedright\arraybackslash}p{(\linewidth - 16\tabcolsep) * \real{0.1056}}
  >{\raggedright\arraybackslash}p{(\linewidth - 16\tabcolsep) * \real{0.1127}}
  >{\raggedright\arraybackslash}p{(\linewidth - 16\tabcolsep) * \real{0.1127}}
  >{\raggedright\arraybackslash}p{(\linewidth - 16\tabcolsep) * \real{0.0915}}@{}}
\caption{{\label{apptbl-effort-ols-sq-inspected}}Linear probability
models of inspection with quadratic position and manipulated reasoning
efforts}\tabularnewline
\toprule\noalign{}
\begin{minipage}[b]{\linewidth}\raggedright
LLM
\end{minipage} & \begin{minipage}[b]{\linewidth}\raggedright
Effort
\end{minipage} & \begin{minipage}[b]{\linewidth}\raggedright
Position
\end{minipage} & \begin{minipage}[b]{\linewidth}\raggedright
Position\(^2\)
\end{minipage} & \begin{minipage}[b]{\linewidth}\raggedright
Price
\end{minipage} & \begin{minipage}[b]{\linewidth}\raggedright
Review score
\end{minipage} & \begin{minipage}[b]{\linewidth}\raggedright
Chain
\end{minipage} & \begin{minipage}[b]{\linewidth}\raggedright
Promotion
\end{minipage} & \begin{minipage}[b]{\linewidth}\raggedright
Adj. \(R^2\)
\end{minipage} \\
\midrule\noalign{}
\endfirsthead
\toprule\noalign{}
\begin{minipage}[b]{\linewidth}\raggedright
LLM
\end{minipage} & \begin{minipage}[b]{\linewidth}\raggedright
Effort
\end{minipage} & \begin{minipage}[b]{\linewidth}\raggedright
Position
\end{minipage} & \begin{minipage}[b]{\linewidth}\raggedright
Position\(^2\)
\end{minipage} & \begin{minipage}[b]{\linewidth}\raggedright
Price
\end{minipage} & \begin{minipage}[b]{\linewidth}\raggedright
Review score
\end{minipage} & \begin{minipage}[b]{\linewidth}\raggedright
Chain
\end{minipage} & \begin{minipage}[b]{\linewidth}\raggedright
Promotion
\end{minipage} & \begin{minipage}[b]{\linewidth}\raggedright
Adj. \(R^2\)
\end{minipage} \\
\midrule\noalign{}
\endhead
\bottomrule\noalign{}
\endlastfoot
Flash Lite & Minimal\textsuperscript{†} & -0.1460*** (0.0115) &
0.0982*** (0.0104) & -0.0001*** (\textless0.0001) & 0.1467*** (0.0014) &
-0.0646*** (0.0013) & -0.0042** (0.0014) & 0.0908 \\
& Low & -0.1474*** (0.0103) & 0.1090*** (0.0091) & -0.0001***
(\textless0.0001) & 0.1026*** (0.0021) & -0.0392*** (0.0013) & 0.0148***
(0.0014) & 0.0650 \\
& Medium & -0.0520*** (0.0090) & 0.0366*** (0.0083) & -0.0001***
(\textless0.0001) & 0.1026*** (0.0023) & -0.0388*** (0.0014) & 0.0163***
(0.0013) & 0.0594 \\
& High & -0.0083 (0.0082) & 0.0079 (0.0079) & -0.0001***
(\textless0.0001) & 0.0854*** (0.0013) & -0.0322*** (0.0010) & 0.0211***
(0.0008) & 0.0523 \\
Flash & Low & -0.0255*** (0.0074) & 0.0163* (0.0069) &
\textgreater-0.0001*** (\textless0.0001) & 0.0654*** (0.0015) &
-0.0225*** (0.0009) & 0.0351*** (0.0008) & 0.0548 \\
& Medium\textsuperscript{†} & -0.1356*** (0.0123) & 0.0993*** (0.0114) &
-0.0001*** (\textless0.0001) & 0.1965*** (0.0033) & -0.0839*** (0.0020)
& -0.0184*** (0.0018) & 0.1090 \\
Pro & Low & -0.0211*** (0.0048) & 0.0170*** (0.0044) &
\textgreater-0.0001*** (\textless0.0001) & 0.0123*** (0.0009) &
-0.0054*** (0.0006) & 0.0053*** (0.0010) & 0.0079 \\
& Medium & -0.0173* (0.0069) & 0.0134* (0.0064) & -0.0001***
(\textless0.0001) & 0.0301*** (0.0016) & -0.0152*** (0.0010) & 0.0095***
(0.0015) & 0.0155 \\
& High\textsuperscript{†} & -0.0567*** (0.0154) & 0.0276 (0.0141) &
-0.0003*** (\textless0.0001) & 0.1282*** (0.0058) & -0.0387*** (0.0023)
& -0.0068** (0.0025) & 0.0345 \\
Sonnet & Low & -0.1292*** (0.0073) & 0.0997*** (0.0064) &
\textgreater-0.0001*** (\textless0.0001) & 0.0369*** (0.0011) &
-0.0146*** (0.0008) & -0.0006 (0.0014) & 0.0319 \\
& High\textsuperscript{†} & -0.0643*** (0.0091) & 0.0454*** (0.0081) &
\textgreater-0.0001*** (\textless0.0001) & 0.0757*** (0.0012) &
-0.0304*** (0.0008) & 0.0265*** (0.0009) & 0.0580 \\
Luna & None & -0.1680*** (0.0143) & 0.1123*** (0.0125) & -0.0001***
(\textless0.0001) & 0.1077*** (0.0016) & -0.0035* (0.0014) & 0.0219***
(0.0013) & 0.0419 \\
& Medium\textsuperscript{†} & -0.1867*** (0.0140) & 0.1272*** (0.0123) &
-0.0001*** (\textless0.0001) & 0.1326*** (0.0020) & -0.0060*** (0.0015)
& 0.0113*** (0.0014) & 0.0479 \\
& xHigh & -0.2137*** (0.0172) & 0.1318*** (0.0154) & -0.0002***
(\textless0.0001) & 0.2036*** (0.0030) & -0.0042* (0.0019) & -0.0099***
(0.0018) & 0.0675 \\
\end{longtable}
}

\emph{Note.} Each model is estimated on 50,000 hotel impressions.
\textsuperscript{†} Provider default reasoning effort. Position is
scaled to {[}0, 1{]} (rank / 100). Standard errors clustered at the
session level in parentheses.

{\def\LTcaptype{apptbl}
\begin{longtable}[]{@{}
  >{\raggedright\arraybackslash}p{(\linewidth - 16\tabcolsep) * \real{0.0915}}
  >{\raggedright\arraybackslash}p{(\linewidth - 16\tabcolsep) * \real{0.0915}}
  >{\raggedright\arraybackslash}p{(\linewidth - 16\tabcolsep) * \real{0.1127}}
  >{\raggedright\arraybackslash}p{(\linewidth - 16\tabcolsep) * \real{0.1056}}
  >{\raggedright\arraybackslash}p{(\linewidth - 16\tabcolsep) * \real{0.1268}}
  >{\raggedright\arraybackslash}p{(\linewidth - 16\tabcolsep) * \real{0.1056}}
  >{\raggedright\arraybackslash}p{(\linewidth - 16\tabcolsep) * \real{0.1127}}
  >{\raggedright\arraybackslash}p{(\linewidth - 16\tabcolsep) * \real{0.1127}}
  >{\raggedright\arraybackslash}p{(\linewidth - 16\tabcolsep) * \real{0.0915}}@{}}
\caption{{\label{apptbl-effort-ols-sq-chosen}}Linear probability models
of choice with quadratic position and manipulated reasoning efforts}\tabularnewline
\toprule\noalign{}
\begin{minipage}[b]{\linewidth}\raggedright
LLM
\end{minipage} & \begin{minipage}[b]{\linewidth}\raggedright
Effort
\end{minipage} & \begin{minipage}[b]{\linewidth}\raggedright
Position
\end{minipage} & \begin{minipage}[b]{\linewidth}\raggedright
Position\(^2\)
\end{minipage} & \begin{minipage}[b]{\linewidth}\raggedright
Price
\end{minipage} & \begin{minipage}[b]{\linewidth}\raggedright
Review score
\end{minipage} & \begin{minipage}[b]{\linewidth}\raggedright
Chain
\end{minipage} & \begin{minipage}[b]{\linewidth}\raggedright
Promotion
\end{minipage} & \begin{minipage}[b]{\linewidth}\raggedright
Adj. \(R^2\)
\end{minipage} \\
\midrule\noalign{}
\endfirsthead
\toprule\noalign{}
\begin{minipage}[b]{\linewidth}\raggedright
LLM
\end{minipage} & \begin{minipage}[b]{\linewidth}\raggedright
Effort
\end{minipage} & \begin{minipage}[b]{\linewidth}\raggedright
Position
\end{minipage} & \begin{minipage}[b]{\linewidth}\raggedright
Position\(^2\)
\end{minipage} & \begin{minipage}[b]{\linewidth}\raggedright
Price
\end{minipage} & \begin{minipage}[b]{\linewidth}\raggedright
Review score
\end{minipage} & \begin{minipage}[b]{\linewidth}\raggedright
Chain
\end{minipage} & \begin{minipage}[b]{\linewidth}\raggedright
Promotion
\end{minipage} & \begin{minipage}[b]{\linewidth}\raggedright
Adj. \(R^2\)
\end{minipage} \\
\midrule\noalign{}
\endhead
\bottomrule\noalign{}
\endlastfoot
Flash Lite & Minimal\textsuperscript{†} & -0.0503*** (0.0067) &
0.0418*** (0.0062) & \textgreater-0.0001*** (\textless0.0001) &
0.0470*** (0.0003) & -0.0185*** (0.0005) & 0.0169*** (0.0016) &
0.0360 \\
& Low & -0.0276*** (0.0066) & 0.0197** (0.0060) & \textgreater-0.0001***
(\textless0.0001) & 0.0449*** (0.0002) & -0.0146*** (0.0004) & 0.0300***
(0.0013) & 0.0433 \\
& Medium & -0.0095 (0.0064) & 0.0052 (0.0060) & \textgreater-0.0001***
(\textless0.0001) & 0.0448*** (0.0002) & -0.0146*** (0.0004) & 0.0331***
(0.0012) & 0.0466 \\
& High & 0.0023 (0.0060) & -0.0010 (0.0059) & \textgreater-0.0001***
(\textless0.0001) & 0.0442*** (0.0001) & -0.0128*** (0.0002) & 0.0421***
(0.0006) & 0.0570 \\
Flash & Low & 0.0034 (0.0058) & -0.0038 (0.0056) &
\textgreater-0.0001*** (\textless0.0001) & 0.0439*** (\textless0.0001) &
-0.0124*** (\textless0.0001) & 0.0445*** (\textless0.0001) & 0.0602 \\
& Medium\textsuperscript{†} & 0.0037 (0.0061) & -0.0051 (0.0059) &
\textgreater-0.0001*** (\textless0.0001) & 0.0439*** (\textless0.0001) &
-0.0124*** (\textless0.0001) & 0.0445*** (\textless0.0001) & 0.0602 \\
Pro & Low & -0.0760*** (0.0075) & 0.0638*** (0.0068) &
\textgreater-0.0001*** (\textless0.0001) & 0.0322*** (0.0010) &
-0.0123*** (0.0009) & 0.0177*** (0.0016) & 0.0240 \\
& Medium & -0.0237*** (0.0068) & 0.0179** (0.0063) &
\textgreater-0.0001*** (\textless0.0001) & 0.0324*** (0.0009) &
-0.0139*** (0.0009) & 0.0176*** (0.0017) & 0.0225 \\
& High\textsuperscript{†} & -0.0080 (0.0065) & 0.0071 (0.0063) &
\textgreater-0.0001*** (\textless0.0001) & 0.0264*** (0.0012) &
-0.0031** (0.0010) & 0.0311*** (0.0015) & 0.0242 \\
Sonnet & Low & -0.1565*** (0.0076) & 0.1217*** (0.0067) &
\textgreater-0.0001*** (\textless0.0001) & 0.0422*** (0.0006) &
-0.0150*** (0.0008) & 0.0025 (0.0015) & 0.0372 \\
& High\textsuperscript{†} & -0.0051 (0.0065) & 0.0016 (0.0061) &
\textgreater-0.0001*** (\textless0.0001) & 0.0439*** (0.0001) &
-0.0128*** (0.0002) & 0.0420*** (0.0006) & 0.0567 \\
Luna & None & -0.0729*** (0.0072) & 0.0544*** (0.0064) &
\textgreater-0.0001*** (\textless0.0001) & 0.0318*** (0.0008) & 0.0007
(0.0009) & 0.0066*** (0.0013) & 0.0137 \\
& Medium\textsuperscript{†} & -0.0652*** (0.0074) & 0.0441*** (0.0065) &
\textgreater-0.0001*** (\textless0.0001) & 0.0273*** (0.0006) &
0.0072*** (0.0008) & -0.0059*** (0.0008) & 0.0132 \\
& xHigh & -0.0364*** (0.0075) & 0.0202** (0.0066) &
\textgreater-0.0001*** (\textless0.0001) & 0.0261*** (0.0004) &
0.0093*** (0.0006) & -0.0023** (0.0008) & 0.0108 \\
\end{longtable}
}

\emph{Note.} Each model is estimated on 50,000 hotel impressions.
\textsuperscript{†} Provider default reasoning effort. Position is
scaled to {[}0, 1{]} (rank / 100). Standard errors clustered at the
session level in parentheses.

{\def\LTcaptype{apptbl}
\begin{longtable}[]{@{}
  >{\raggedright\arraybackslash}p{(\linewidth - 10\tabcolsep) * \real{0.3458}}
  >{\raggedright\arraybackslash}p{(\linewidth - 10\tabcolsep) * \real{0.1215}}
  >{\raggedright\arraybackslash}p{(\linewidth - 10\tabcolsep) * \real{0.1308}}
  >{\raggedright\arraybackslash}p{(\linewidth - 10\tabcolsep) * \real{0.1215}}
  >{\raggedright\arraybackslash}p{(\linewidth - 10\tabcolsep) * \real{0.1215}}
  >{\raggedright\arraybackslash}p{(\linewidth - 10\tabcolsep) * \real{0.1215}}@{}}
\caption{{\label{apptbl-robustness-descriptive}}Search and choice
behavior under the robustness conditions}\tabularnewline
\toprule\noalign{}
\multirow{2}{=}{\begin{minipage}[b]{\linewidth}\raggedright
\end{minipage}} &
\multicolumn{2}{>{\raggedright\arraybackslash}p{(\linewidth - 10\tabcolsep) * \real{0.2523} + 2\tabcolsep}}{%
\begin{minipage}[b]{\linewidth}\raggedright
Flash Lite
\end{minipage}} &
\multicolumn{3}{>{\raggedright\arraybackslash}p{(\linewidth - 10\tabcolsep) * \real{0.3645} + 4\tabcolsep}@{}}{%
\begin{minipage}[b]{\linewidth}\raggedright
Flash
\end{minipage}} \\
& \begin{minipage}[b]{\linewidth}\raggedright
Main
\end{minipage} & \begin{minipage}[b]{\linewidth}\raggedright
Alternative
\end{minipage} & \begin{minipage}[b]{\linewidth}\raggedright
Main
\end{minipage} & \begin{minipage}[b]{\linewidth}\raggedright
No name
\end{minipage} & \begin{minipage}[b]{\linewidth}\raggedright
Finish
\end{minipage} \\
\midrule\noalign{}
\endfirsthead
\toprule\noalign{}
\multirow{2}{=}{\begin{minipage}[b]{\linewidth}\raggedright
\end{minipage}} &
\multicolumn{2}{>{\raggedright\arraybackslash}p{(\linewidth - 10\tabcolsep) * \real{0.2523} + 2\tabcolsep}}{%
\begin{minipage}[b]{\linewidth}\raggedright
Flash Lite
\end{minipage}} &
\multicolumn{3}{>{\raggedright\arraybackslash}p{(\linewidth - 10\tabcolsep) * \real{0.3645} + 4\tabcolsep}@{}}{%
\begin{minipage}[b]{\linewidth}\raggedright
Flash
\end{minipage}} \\
& \begin{minipage}[b]{\linewidth}\raggedright
Main
\end{minipage} & \begin{minipage}[b]{\linewidth}\raggedright
Alternative
\end{minipage} & \begin{minipage}[b]{\linewidth}\raggedright
Main
\end{minipage} & \begin{minipage}[b]{\linewidth}\raggedright
No name
\end{minipage} & \begin{minipage}[b]{\linewidth}\raggedright
Finish
\end{minipage} \\
\midrule\noalign{}
\endhead
\bottomrule\noalign{}
\endlastfoot
\emph{N} & 500 & 500 & 500 & 500 & 500 \\
Conversion rate & 100.0\% & 100.0\% & 100.0\% & 100.0\% & 100.0\% \\
Outside option selected & 0.0\% & 0.0\% & 0.0\% & 0.0\% & 0.0\% \\
Inspections per session & & & & & \\
(\emph{Mean}) & 3.12 & 2.92 & 4.25 & 4.73 & 3.63 \\
(\emph{Median}) & 3.00 & 3.00 & 4.00 & 4.50 & 3.00 \\
(\emph{Mode}) & 3.00 & 3.00 & 4.00 & 4.00 & 3.00 \\
(\emph{SD}) & 0.54 & 0.72 & 1.59 & 1.58 & 1.47 \\
Sessions with one inspection & 0.8\% & 6.2\% & 2.6\% & 0.0\% & 5.6\% \\
Chose the first inspected hotel & 62.0\% & 65.6\% & 99.8\% & 93.8\% &
99.8\% \\
Position of the chosen hotel (\emph{M}) & 43.97 & 42.58 & 49.46 & 49.56
& 53.12 \\
Modal choice proportion & 61.4\% & 60.8\% & 100.0\% & 93.8\% &
100.0\% \\
\end{longtable}
}

\emph{Note.} Main = main prompt; Alternative = prompt without the
stopping rule and the recommendation question; No name = hotel names
removed from the listing page; Finish = neutral \emph{finish} tool with
an optional hotel and no booking instruction. Flash Lite uses minimal
and Flash medium reasoning effort, the provider defaults.

{\def\LTcaptype{apptbl}
\begin{longtable}[]{@{}
  >{\raggedright\arraybackslash}p{(\linewidth - 10\tabcolsep) * \real{0.1966}}
  >{\raggedright\arraybackslash}p{(\linewidth - 10\tabcolsep) * \real{0.1538}}
  >{\raggedright\arraybackslash}p{(\linewidth - 10\tabcolsep) * \real{0.1538}}
  >{\raggedright\arraybackslash}p{(\linewidth - 10\tabcolsep) * \real{0.1538}}
  >{\raggedright\arraybackslash}p{(\linewidth - 10\tabcolsep) * \real{0.1538}}
  >{\raggedright\arraybackslash}p{(\linewidth - 10\tabcolsep) * \real{0.1538}}@{}}
\caption{{\label{apptbl-robustness-ols}}Linear probability models of
inspection and choice under the robustness conditions}\tabularnewline
\toprule\noalign{}
\multirow{2}{=}{\begin{minipage}[b]{\linewidth}\raggedright
\end{minipage}} &
\multicolumn{2}{>{\raggedright\arraybackslash}p{(\linewidth - 10\tabcolsep) * \real{0.3077} + 2\tabcolsep}}{%
\begin{minipage}[b]{\linewidth}\raggedright
Flash Lite
\end{minipage}} &
\multicolumn{3}{>{\raggedright\arraybackslash}p{(\linewidth - 10\tabcolsep) * \real{0.4615} + 4\tabcolsep}@{}}{%
\begin{minipage}[b]{\linewidth}\raggedright
Flash
\end{minipage}} \\
& \begin{minipage}[b]{\linewidth}\raggedright
Main
\end{minipage} & \begin{minipage}[b]{\linewidth}\raggedright
Alternative
\end{minipage} & \begin{minipage}[b]{\linewidth}\raggedright
Main
\end{minipage} & \begin{minipage}[b]{\linewidth}\raggedright
No name
\end{minipage} & \begin{minipage}[b]{\linewidth}\raggedright
Finish
\end{minipage} \\
\midrule\noalign{}
\endfirsthead
\toprule\noalign{}
\multirow{2}{=}{\begin{minipage}[b]{\linewidth}\raggedright
\end{minipage}} &
\multicolumn{2}{>{\raggedright\arraybackslash}p{(\linewidth - 10\tabcolsep) * \real{0.3077} + 2\tabcolsep}}{%
\begin{minipage}[b]{\linewidth}\raggedright
Flash Lite
\end{minipage}} &
\multicolumn{3}{>{\raggedright\arraybackslash}p{(\linewidth - 10\tabcolsep) * \real{0.4615} + 4\tabcolsep}@{}}{%
\begin{minipage}[b]{\linewidth}\raggedright
Flash
\end{minipage}} \\
& \begin{minipage}[b]{\linewidth}\raggedright
Main
\end{minipage} & \begin{minipage}[b]{\linewidth}\raggedright
Alternative
\end{minipage} & \begin{minipage}[b]{\linewidth}\raggedright
Main
\end{minipage} & \begin{minipage}[b]{\linewidth}\raggedright
No name
\end{minipage} & \begin{minipage}[b]{\linewidth}\raggedright
Finish
\end{minipage} \\
\midrule\noalign{}
\endhead
\bottomrule\noalign{}
\endlastfoot
\emph{Panel A: Inspected} & & & & & \\
Position & -0.1460*** (0.0115) & -0.1514*** (0.0108) & -0.1356***
(0.0123) & -0.0816*** (0.0139) & -0.1308*** (0.0125) \\
Position\(^2\) & 0.0982*** (0.0104) & 0.1086*** (0.0097) & 0.0993***
(0.0114) & 0.0582*** (0.0128) & 0.1022*** (0.0112) \\
Price & -0.0001*** (\textless0.0001) & -0.0001*** (\textless0.0001) &
-0.0001*** (\textless0.0001) & -0.0001*** (\textless0.0001) & -0.0001***
(\textless0.0001) \\
Review score & 0.1467*** (0.0014) & 0.1379*** (0.0017) & 0.1965***
(0.0033) & 0.2218*** (0.0031) & 0.1665*** (0.0030) \\
Chain & -0.0646*** (0.0013) & -0.0602*** (0.0013) & -0.0839*** (0.0020)
& -0.1085*** (0.0018) & -0.0679*** (0.0018) \\
Promotion & -0.0042** (0.0014) & -0.0016 (0.0013) & -0.0184*** (0.0018)
& -0.0500*** (0.0017) & -0.0085*** (0.0017) \\
Adj. \(R^2\) & 0.0908 & 0.0860 & 0.1090 & 0.1301 & 0.0897 \\
N & 50,000 & 50,000 & 50,000 & 50,000 & 50,000 \\
\emph{Panel B: Chosen} & & & & & \\
Position & -0.0503*** (0.0067) & -0.0626*** (0.0070) & 0.0037 (0.0061) &
0.0098 (0.0062) & -0.0002 (0.0059) \\
Position\(^2\) & 0.0418*** (0.0062) & 0.0524*** (0.0064) & -0.0051
(0.0059) & -0.0108 (0.0058) & 0.0025 (0.0056) \\
Price & \textgreater-0.0001*** (\textless0.0001) &
\textgreater-0.0001*** (\textless0.0001) & \textgreater-0.0001***
(\textless0.0001) & \textgreater-0.0001*** (\textless0.0001) &
\textgreater-0.0001*** (\textless0.0001) \\
Review score & 0.0470*** (0.0003) & 0.0467*** (0.0003) & 0.0439***
(\textless0.0001) & 0.0446*** (0.0001) & 0.0439*** (\textless0.0001) \\
Chain & -0.0185*** (0.0005) & -0.0179*** (0.0005) & -0.0124***
(\textless0.0001) & -0.0137*** (0.0002) & -0.0124***
(\textless0.0001) \\
Promotion & 0.0169*** (0.0016) & 0.0168*** (0.0016) & 0.0445***
(\textless0.0001) & 0.0399*** (0.0008) & 0.0445*** (\textless0.0001) \\
Adj. \(R^2\) & 0.0360 & 0.0360 & 0.0602 & 0.0550 & 0.0602 \\
N & 50,000 & 50,000 & 50,000 & 50,000 & 50,000 \\
\end{longtable}
}

\emph{Note.} Position is scaled to {[}0, 1{]} (rank / 100). Standard
errors clustered at the session level in parentheses. Chain is derived
from the hotel name, which is not displayed in the No name condition.
Flash Lite uses minimal and Flash medium reasoning effort, the provider
defaults. *p\textless0.05, **p\textless0.01, ***p\textless0.001

{\def\LTcaptype{apptbl}
\begin{longtable}[]{@{}
  >{\raggedright\arraybackslash}p{(\linewidth - 8\tabcolsep) * \real{0.4126}}
  >{\raggedright\arraybackslash}p{(\linewidth - 8\tabcolsep) * \real{0.1049}}
  >{\raggedright\arraybackslash}p{(\linewidth - 8\tabcolsep) * \real{0.1399}}
  >{\raggedright\arraybackslash}p{(\linewidth - 8\tabcolsep) * \real{0.2378}}
  >{\raggedright\arraybackslash}p{(\linewidth - 8\tabcolsep) * \real{0.1049}}@{}}
\caption{{\label{apptbl-open-descriptive}}Descriptive statistics for
open-weight LLMs tested on the same setup as the main experiment}\tabularnewline
\toprule\noalign{}
\begin{minipage}[b]{\linewidth}\raggedright
\end{minipage} & \begin{minipage}[b]{\linewidth}\raggedright
Qwen3.8 27B
\end{minipage} & \begin{minipage}[b]{\linewidth}\raggedright
Muse Glimmer 30B
\end{minipage} & \begin{minipage}[b]{\linewidth}\raggedright
Nemotron 3.5 Lightning 30B A3B
\end{minipage} & \begin{minipage}[b]{\linewidth}\raggedright
Gemma 4 31B
\end{minipage} \\
\midrule\noalign{}
\endfirsthead
\toprule\noalign{}
\begin{minipage}[b]{\linewidth}\raggedright
\end{minipage} & \begin{minipage}[b]{\linewidth}\raggedright
Qwen3.8 27B
\end{minipage} & \begin{minipage}[b]{\linewidth}\raggedright
Muse Glimmer 30B
\end{minipage} & \begin{minipage}[b]{\linewidth}\raggedright
Nemotron 3.5 Lightning 30B A3B
\end{minipage} & \begin{minipage}[b]{\linewidth}\raggedright
Gemma 4 31B
\end{minipage} \\
\midrule\noalign{}
\endhead
\bottomrule\noalign{}
\endlastfoot
N & 500 & 500 & 500 & 500 \\
Conversion rate & 100.0\% & 100.0\% & 97.8\% & 100.0\% \\
Outside option selected & 0.0\% & 0.0\% & 2.2\% & 0.0\% \\
Inspections per session & & & & \\
\emph{(Mean)} & 4.19 & 9.62 & 1.76 & 3.99 \\
\emph{(Median)} & 4.00 & 10.00 & 1.00 & 4.00 \\
\emph{(Mode)} & 4.00 & 9.00 & 1.00 & 4.00 \\
\emph{(SD)} & 1.32 & 2.27 & 1.56 & 1.39 \\
Sessions with one inspection & 0.0\% & 0.0\% & 27.4\% & 0.2\% \\
Chose the first inspected hotel & 87.8\% & 62.6\% & 62.4\% & 76.6\% \\
Position (rank) of the chosen hotel (M) & 51.83 & 39.89 & 47.56 &
52.09 \\
Modal choice proportion & 88.2\% & 62.4\% & 67.3\% & 76.6\% \\
\end{longtable}
}

{\def\LTcaptype{apptbl}
\begin{longtable}[]{@{}
  >{\raggedright\arraybackslash}p{(\linewidth - 14\tabcolsep) * \real{0.2230}}
  >{\raggedright\arraybackslash}p{(\linewidth - 14\tabcolsep) * \real{0.1081}}
  >{\raggedright\arraybackslash}p{(\linewidth - 14\tabcolsep) * \real{0.1014}}
  >{\raggedright\arraybackslash}p{(\linewidth - 14\tabcolsep) * \real{0.1216}}
  >{\raggedright\arraybackslash}p{(\linewidth - 14\tabcolsep) * \real{0.1014}}
  >{\raggedright\arraybackslash}p{(\linewidth - 14\tabcolsep) * \real{0.1081}}
  >{\raggedright\arraybackslash}p{(\linewidth - 14\tabcolsep) * \real{0.1081}}
  >{\raggedright\arraybackslash}p{(\linewidth - 14\tabcolsep) * \real{0.0878}}@{}}
\caption{{\label{apptbl-open-ols}}Linear probability models of
inspection and choice with quadratic position for open-weight LLMs}\tabularnewline
\toprule\noalign{}
\begin{minipage}[b]{\linewidth}\raggedright
LLM
\end{minipage} & \begin{minipage}[b]{\linewidth}\raggedright
Position
\end{minipage} & \begin{minipage}[b]{\linewidth}\raggedright
Position\(^2\)
\end{minipage} & \begin{minipage}[b]{\linewidth}\raggedright
Price
\end{minipage} & \begin{minipage}[b]{\linewidth}\raggedright
Review score
\end{minipage} & \begin{minipage}[b]{\linewidth}\raggedright
Chain
\end{minipage} & \begin{minipage}[b]{\linewidth}\raggedright
Promotion
\end{minipage} & \begin{minipage}[b]{\linewidth}\raggedright
Adj. \(R^2\)
\end{minipage} \\
\midrule\noalign{}
\endfirsthead
\toprule\noalign{}
\begin{minipage}[b]{\linewidth}\raggedright
LLM
\end{minipage} & \begin{minipage}[b]{\linewidth}\raggedright
Position
\end{minipage} & \begin{minipage}[b]{\linewidth}\raggedright
Position\(^2\)
\end{minipage} & \begin{minipage}[b]{\linewidth}\raggedright
Price
\end{minipage} & \begin{minipage}[b]{\linewidth}\raggedright
Review score
\end{minipage} & \begin{minipage}[b]{\linewidth}\raggedright
Chain
\end{minipage} & \begin{minipage}[b]{\linewidth}\raggedright
Promotion
\end{minipage} & \begin{minipage}[b]{\linewidth}\raggedright
Adj. \(R^2\)
\end{minipage} \\
\midrule\noalign{}
\endhead
\bottomrule\noalign{}
\endlastfoot
\emph{Panel A: Inspected} & & & & & & & \\
Gemma 4 31B & -0.0725*** (0.0120) & 0.0562*** (0.0113) & -0.0001***
(\textless0.0001) & 0.1638*** (0.0021) & -0.0485*** (0.0012) &
-0.0152*** (0.0011) & 0.0654 \\
Muse Glimmer 30B & -0.4536*** (0.0200) & 0.2781*** (0.0178) & -0.0004***
(\textless0.0001) & 0.2302*** (0.0031) & -0.0313*** (0.0022) &
-0.0242*** (0.0025) & 0.0886 \\
Nemotron 3.5 Lightning 30B A3B & -0.0436*** (0.0107) & 0.0295** (0.0093)
& -0.0001*** (\textless0.0001) & 0.0531*** (0.0022) & -0.0180***
(0.0013) & 0.0127*** (0.0015) & 0.0224 \\
Qwen3.8 27B & -0.0356** (0.0126) & 0.0136 (0.0116) & -0.0001***
(\textless0.0001) & 0.1689*** (0.0022) & -0.0391*** (0.0016) &
-0.0130*** (0.0013) & 0.0619 \\
\emph{Panel B: Chosen} & & & & & & & \\
Gemma 4 31B & -0.0048 (0.0062) & 0.0067 (0.0060) &
\textgreater-0.0001*** (\textless0.0001) & 0.0450*** (0.0003) &
-0.0149*** (0.0005) & 0.0281*** (0.0013) & 0.0409 \\
Muse Glimmer 30B & -0.0345*** (0.0069) & 0.0216*** (0.0062) &
\textgreater-0.0001*** (\textless0.0001) & 0.0377*** (0.0009) &
-0.0116*** (0.0008) & 0.0262*** (0.0015) & 0.0323 \\
Nemotron 3.5 Lightning 30B A3B & -0.0062 (0.0066) & 0.0030 (0.0061) &
\textgreater-0.0001*** (\textless0.0001) & 0.0362*** (0.0007) &
-0.0119*** (0.0007) & 0.0258*** (0.0014) & 0.0304 \\
Qwen3.8 27B & 0.0124* (0.0060) & -0.0108 (0.0057) &
\textgreater-0.0001*** (\textless0.0001) & 0.0449*** (0.0002) &
-0.0145*** (0.0003) & 0.0359*** (0.0011) & 0.0503 \\
\end{longtable}
}

\emph{Note.} Each model is estimated on 50,000 hotel impressions.
Position is scaled to {[}0, 1{]} (rank / 100). Standard errors clustered
at the session level in parentheses.\\
*p\textless0.05, **p\textless0.01, ***p\textless0.001

\end{landscape}

{\def\LTcaptype{apptbl}
\begin{longtable}[]{@{}
  >{\raggedright\arraybackslash}p{(\linewidth - 10\tabcolsep) * \real{0.1327}}
  >{\raggedright\arraybackslash}p{(\linewidth - 10\tabcolsep) * \real{0.1327}}
  >{\raggedright\arraybackslash}p{(\linewidth - 10\tabcolsep) * \real{0.2347}}
  >{\raggedright\arraybackslash}p{(\linewidth - 10\tabcolsep) * \real{0.1020}}
  >{\raggedright\arraybackslash}p{(\linewidth - 10\tabcolsep) * \real{0.1122}}
  >{\raggedright\arraybackslash}p{(\linewidth - 10\tabcolsep) * \real{0.2449}}@{}}
\caption{{\label{apptbl-effort-cost}}Position dependence of choice,
token usage, and cost per session by LLM and reasoning effort}\tabularnewline
\toprule\noalign{}
\multirow{2}{=}{\begin{minipage}[b]{\linewidth}\raggedright
LLM
\end{minipage}} &
\multirow{2}{=}{\begin{minipage}[b]{\linewidth}\raggedright
Effort
\end{minipage}} &
\multirow{2}{=}{\begin{minipage}[b]{\linewidth}\raggedright
Position \(\chi^2\)(2)
\end{minipage}} &
\multicolumn{2}{>{\raggedright\arraybackslash}p{(\linewidth - 10\tabcolsep) * \real{0.2143} + 2\tabcolsep}}{%
\begin{minipage}[b]{\linewidth}\raggedright
Tokens per session
\end{minipage}} &
\multirow{2}{=}{\begin{minipage}[b]{\linewidth}\raggedright
Cost per session (\$)
\end{minipage}} \\
& & & \begin{minipage}[b]{\linewidth}\raggedright
Input
\end{minipage} & \begin{minipage}[b]{\linewidth}\raggedright
Output
\end{minipage} \\
\midrule\noalign{}
\endfirsthead
\toprule\noalign{}
\multirow{2}{=}{\begin{minipage}[b]{\linewidth}\raggedright
LLM
\end{minipage}} &
\multirow{2}{=}{\begin{minipage}[b]{\linewidth}\raggedright
Effort
\end{minipage}} &
\multirow{2}{=}{\begin{minipage}[b]{\linewidth}\raggedright
Position \(\chi^2\)(2)
\end{minipage}} &
\multicolumn{2}{>{\raggedright\arraybackslash}p{(\linewidth - 10\tabcolsep) * \real{0.2143} + 2\tabcolsep}}{%
\begin{minipage}[b]{\linewidth}\raggedright
Tokens per session
\end{minipage}} &
\multirow{2}{=}{\begin{minipage}[b]{\linewidth}\raggedright
Cost per session (\$)
\end{minipage}} \\
& & & \begin{minipage}[b]{\linewidth}\raggedright
Input
\end{minipage} & \begin{minipage}[b]{\linewidth}\raggedright
Output
\end{minipage} \\
\midrule\noalign{}
\endhead
\bottomrule\noalign{}
\endlastfoot
Flash Lite & Minimal\textsuperscript{†} & 60.36*** & 52,282 & 80 &
0.0132 \\
& Low & 29.90*** & 42,536 & 479 & 0.0114 \\
& Medium & 8.66* & 43,649 & 1,041 & 0.0125 \\
& High & 0.68 & 38,940 & 1,065 & 0.0113 \\
Flash & Low & 0.49 & 30,403 & 339 & 0.0481 \\
& Medium\textsuperscript{†} & 1.68 & 75,163 & 874 & 0.1193 \\
Pro & Low & 107.21*** & 16,752 & 293 & 0.0370 \\
& Medium & 17.12*** & 27,557 & 590 & 0.0622 \\
& High\textsuperscript{†} & 1.56 & 125,960 & 2,552 & 0.2825 \\
Sonnet & Low & 609.27*** & 29,992 & 233 & 0.0623 \\
& High\textsuperscript{†} & 5.56 & 43,860 & 787 & 0.0956 \\
Luna & None & 145.58*** & 50,277 & 91 & 0.0051 \\
& Medium\textsuperscript{†} & 187.49*** & 57,665 & 571 & 0.0061 \\
& xHigh & 120.06*** & 91,640 & 1,592 & 0.0100 \\
\end{longtable}
}

\emph{Note.} Position \(\chi^2\)(2) is the Wald statistic for the joint
hypothesis that the coefficients on position and position\(^2\) in the
model of choice are zero
(\quartoapptblref{apptbl-effort-ols-sq-chosen}); a non-significant
statistic indicates no detectable dependence of choice on position.
Token counts and costs are means over 500 sessions per condition. Output
tokens include reasoning tokens, which providers bill at the output
rate. Cost is computed from token counts at each provider's list price
per million tokens. \textsuperscript{†} Provider default reasoning
effort.\\
*p\textless0.05, **p\textless0.01, ***p\textless0.001

\section*{Web Appendix B - Supplementary figures}\label{sec-appendix-b}
\addcontentsline{toc}{section}{Web Appendix B - Supplementary figures}

\begin{appfig}

\centering{

\pandocbounded{\includegraphics[keepaspectratio]{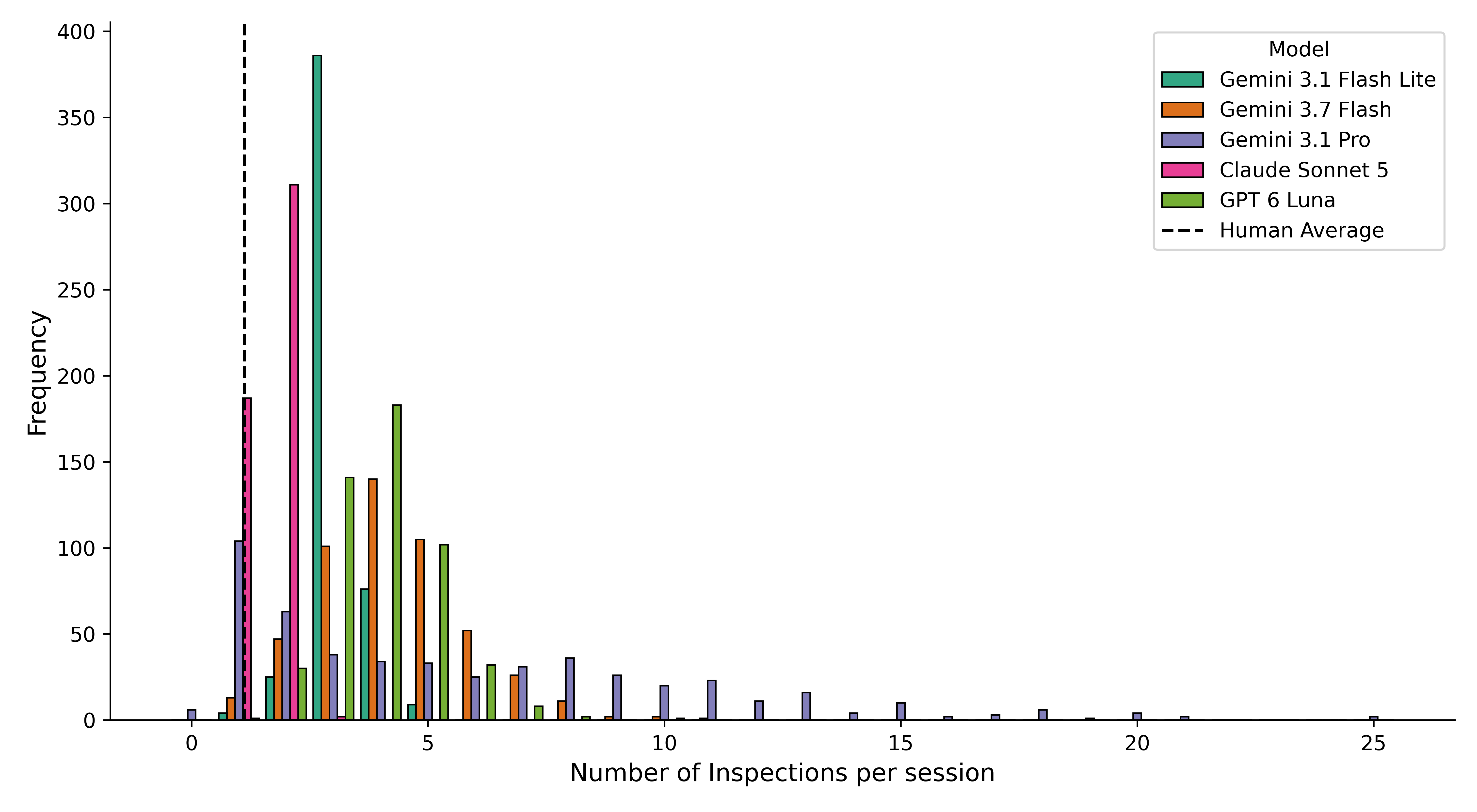}}

}

\caption{\label{appfig-insp-hist}Distribution of inspections per session
by model}

\floatnote{\emph{Note.} Vertical line denotes the human benchmark mean \citep[1.12
clicks per impression;][]{ursu2018power}.}

\end{appfig}%

\section*{Web Appendix C - Construction of the simulated search
environment}\label{sec-appendix-c}
\addcontentsline{toc}{section}{Web Appendix C - Construction of the
simulated search environment}

To provide an empirical test of LLM decision making against human
benchmark data, we constructed a simulated hotel choice environment
calibrated to the randomized ranking experiment in
\citet{ursu2018power}. This section details the empirical calibration of
the search parameters, choice set extraction, attribute partitioning,
and econometric specifications used to replicate the reduced-form
benchmarks.

\textbf{Search query.} We calibrated the simulated search parameters to
match the median search impression characteristics reported in the human
benchmark data \citep[Table 1]{ursu2018power}. In the original Expedia
dataset, the median search impression involved a two-day trip length, an
18-day advance booking window, a Saturday night stay, two adults, zero
children, and one room. To operationalize these parameters with
contemporary search data, the query was executed on Expedia, on May 13,
2026, specifying a check-in date of Saturday, May 30, 2026, and a
check-out date of Monday, June 1, 2026, for two adult guests sharing a
single room.

\textbf{Destination.} In the human benchmark study, the primary Expedia
dataset anonymized destination cities while restricting structural
estimation to the four largest destination markets in the United States,
which accounted for approximately 80\% of domestic queries. To validate
sequential search behavior and click-order dynamics,
\citet{ursu2018power} utilized a companion dataset from the Wharton
Customer Analytics Initiative that tracked hotel searches in Manhattan,
New York. Following this empirical precedent, we restricted our search
query to the Manhattan market to ensure comparability with the human
benchmark.

\textbf{Listing page.} Following the observational boundary established
in the human benchmark \citep[Section 3.2.1]{ursu2018power}, data
collection was restricted to the first search engine results page prior
to any user pagination. The live query yielded a choice set of 100 hotel
listings. For each hotel, all information visible on the initial results
page was extracted and structured into the pre-click choice set,
alongside the hidden post-click attributes retrieved through the
\emph{inspect} tool.

\textbf{Information architecture.} In the human benchmark, consumers
evaluated hotels in two stages, first observing a listing page and then
clicking through to a hotel page with detailed information
\citep{ursu2018power}. To preserve this structure, hotel information was
partitioned into pre-click listing-page attributes and post-click
detailed attributes. The pre-click layer, loaded into the context window
of the agent as a single block of text, contained the hotel name,
neighborhood, aggregate guest review rating, total review count, nightly
price, total stay price, original price, promotional discount badge
presence, rooms-left scarcity alerts, and promotional status badges
(e.g., VIP or sponsored status). The post-click layer, which corresponds
to the hotel page that a human consumer reaches by clicking, was
accessible through the \emph{inspect} tool and contained detailed
property attributes, including the star rating, sub-ratings for
cleanliness, service, amenities, and location, detailed room
configurations, and cancellation policies. Ending a session without a
booking corresponds to a human consumer leaving the results page without
booking and therefore represents the outside option.

\textbf{Econometric specification.} To make the LLM regression models
comparable to the human ones, we estimated linear probability models
corresponding to the reduced-form specifications in
\citet{ursu2018power}, Table 2. The regression models included five
primary pre-click covariates: listing position (indexed 1 to 100),
nightly price in dollars, guest review score (normalized from a 10-point
to a 5-point scale), a chain proxy, and a promotion proxy. The chain
proxy was coded as a binary indicator for major national or
international hotel brands identified through keyword matching against
property titles (e.g., Marriott, Hilton, Hyatt, Sheraton, Holiday Inn,
Westin, DoubleTree, Courtyard, Fairfield, Kimpton, Wyndham, Sofitel,
Aloft, Embassy Suites, and InterContinental). The promotion proxy was
coded as a binary indicator for the presence of a promotional discount
badge on the listing page. To capture non-monotonic position effects, we
additionally estimated specifications that include the square of
position, with position rescaled to the unit interval.

\textbf{Omitted variables.} The empirical replication remains an
approximation of the original field data due to structural evolutions in
the Expedia interface. Two control variables from the human benchmark
were omitted from the pre-click specification: the proprietary 0 to 7
location score and the listing-page star rating. These variables were
not displayed on the modern Expedia listing page prior to inspection and
were therefore excluded from the pre-click control set to ensure that
the empirical model controls strictly for information observable to the
agent before an inspection decision is made. Furthermore, post-click
cues (e.g., specific room amenities and detailed sub-ratings) were
excluded from the pre-click linear probability models because they are
revealed only after inspection tool invocation.

\section*{Web Appendix D - Large language model specifications and
runtime configurations}\label{sec-appendix-d}
\addcontentsline{toc}{section}{Web Appendix D - Large language model
specifications and runtime configurations}

This section details the model endpoints, default reasoning effort
tiers, sampling parameters, and tool invocation protocols across all
experimental conditions. LLM interactions were managed through the
official Google GenAI Python SDK, Anthropic Python SDK, and OpenAI
Python SDK, and the open-weight LLMs were served locally through
llama.cpp.

\textbf{Model endpoints and versions.} Google LLMs were initialized
using their official API identifiers: gemini-3.1-flash-lite,
gemini-3.7-flash, and gemini-3.1-pro-preview. The Anthropic model was
deployed using the claude-sonnet-5 endpoint with a maximum output limit
of 10,000 tokens to accommodate extended internal reasoning and tool
execution outputs. The OpenAI model was deployed using the gpt-6-luna
endpoint through the Responses API, with the provider-default output
limit.

\textbf{Open-weight models.} The four open-weight LLMs were run locally
from Unsloth Dynamic GGUF quantizations: Gemma-4-31B-it (Q4\_K\_XL),
Muse-Glimmer-30B (Q6\_K\_XL), NVIDIA-Nemotron-3.5-Lightning-30B-A3B
(Q4\_K\_XL), and Qwen3.8-27B (Q6\_K\_XL). Each LLM was served by
llama-server with all layers offloaded to the GPU and a single inference
slot, so that sessions were processed one at a time, and was queried
through the OpenAI-compatible Responses API of the server. No reasoning
effort was specified for these LLMs, which therefore operated under the
default reasoning behavior of their chat templates.

\textbf{Default reasoning effort.} In the baseline out-of-the-box
condition, LLMs were evaluated under their provider-default reasoning
effort settings. Gemini 3.1 Flash Lite defaults to minimal; Gemini 3.7
Flash defaults to medium; and Gemini 3.1 Pro defaults to high reasoning
effort. For Anthropic, Claude Sonnet 5 was deployed using adaptive
thinking with high effort. For OpenAI, GPT 6 Luna defaults to medium
reasoning effort.

\textbf{Reasoning effort manipulations.} In the reasoning effort
experiment, for Gemini 3.1 Flash Lite, we evaluated all four available
effort tiers: minimal, low, medium, and high. For Gemini 3.7 Flash, we
evaluated low and medium effort. For Gemini 3.1 Pro, we evaluated all
three available effort tiers: low, medium, and high. For Claude Sonnet
5, we evaluated low and high effort. For GPT 6 Luna, we evaluated none,
medium, and xHigh effort.

\textbf{Sampling parameter handling.} Contemporary reasoning LLMs
enforce strict constraints on sampling parameters when internal
deliberation is active. For Google LLMs accessed via the Google GenAI
SDK, the provider default parameters are temperature of 1.0, top-p of
0.95, and top-k of 64. For Anthropic LLMs accessed via the Anthropic
SDK, the default parameters are temperature of 1.0 and top-p of 1.0.
When extended or adaptive thinking is enabled, both Google and Anthropic
API endpoints lock temperature strictly to 1.0 and reject custom
temperature or sampling overrides with client errors. For the OpenAI
LLM, no sampling parameters were specified, and requests used the
provider defaults. Likewise, no sampling parameters were specified for
the open-weight LLMs, which were therefore sampled under the default
settings of llama-server. As a result, all LLMs were evaluated under
their native out-of-the-box sampling distributions.

\textbf{Tool execution constraints.} To enforce the sequential search,
LLMs were restricted to executing a single tool per conversational turn.
Parallel tool execution was disabled at the API level
(\emph{disable\_parallel\_tool\_use} set to \emph{true} for Anthropic
and \emph{parallel\_tool\_calls} set to \emph{false} for OpenAI and the
open-weight LLMs) or through single-tool constraints specified in system
instructions for Google. Each session was executed as an independent
replication with a newly initialized context window to prevent state
leakage or cross-session memory carryover.

\end{document}